\documentclass[preprint,10pt, authoryear]{elsarticle}

\usepackage{graphicx}%
\usepackage{multirow}%
\usepackage{amsmath,amssymb,amsfonts}%
\usepackage{amsthm}%
\usepackage{mathrsfs}%
\usepackage[title]{appendix}%
\usepackage{xcolor}%
\usepackage{textcomp}%
\usepackage{manyfoot}%
\usepackage{booktabs}%
\usepackage{algorithm}%
\usepackage{algorithmicx}%
\usepackage{algpseudocode}%
\usepackage{listings}%
\usepackage{colortbl}
\usepackage{xurl}
\usepackage{tabularx}
\usepackage{hyperref}

\definecolor{codegreen}{rgb}{0,0.6,0}
\definecolor{codegray}{rgb}{0.5,0.5,0.5}
\definecolor{codepurple}{rgb}{0.58,0,0.82}
\definecolor{backcolour}{rgb}{0.95,0.95,0.92}

\begin{document}

\makeatletter
\def\ps@pprintTitle{%
    \let\@oddhead\@empty
    \let\@evenhead\@empty
    \def\@oddfoot{}%
    \def\@evenfoot{}%
}
\makeatother

\begin{frontmatter}

\title{Advancing the Biological Plausibility and Efficacy of Hebbian Convolutional Neural Networks}

\author[1]{Julian Jiménez Nimmo\fnref{fn1}}

\author[1]{Esther Mondragón\corref{cor1}}
\ead{e.mondragon@citystgeorges.ac.uk}

\cortext[cor1]{Corresponding author}
\fntext[fn1]{First author.}

\affiliation[1]{organization={Artificial Intelligence Research Centre (CitAI), Department of Computer Science, City St George's, University of London},
addressline={Northampton Square},
postcode={EC1V 0HBC},
city={London},
country={United Kingdom}}

\begin{abstract}
The research presented in this paper advances the integration of Hebbian learning into Convolutional Neural Networks (CNNs) for image processing, systematically exploring different architectures to build an optimal configuration, adhering to biological tenability. Hebbian learning operates on local unsupervised neural information to form feature representations, providing an alternative to the popular but arguably biologically implausible and computationally intensive backpropagation learning algorithm. The suggested optimal architecture significantly enhances recent research aimed at integrating Hebbian learning with competition mechanisms and CNNs, expanding their representational capabilities by incorporating hard Winner-Takes-All (WTA) competition, Gaussian lateral inhibition mechanisms and Bienenstock–Cooper–Munro (BCM) learning rule in a single model. Mean accuracy classification measures during the last half of test epochs on CIFAR-10 revealed that the resulting optimal model matched its end-to-end backpropagation variant with 75.2\% each, critically surpassing the state-of-the-art hard-WTA performance in CNNs of the same network depth (64.6\%) by 10.6\%. It also achieved competitive performance on MNIST (98\%) and STL-10 (69.5\%). Moreover, results showed clear indications of sparse hierarchical learning through increasingly complex and abstract receptive fields. In summary, our implementation enhances both the performance and the generalisability of the learnt representations and constitutes a crucial step towards more biologically realistic artificial neural networks
\end{abstract}

\begin{keyword}
Hebbian Learning \sep Unsupervised Feature Learning
 \sep Convolutional Neural Networks (CNNs)
 \sep Sparse Neural Coding \sep Hard/Soft Competition

\end{keyword}

\end{frontmatter}

\section{Introduction}

Deep neural networks (DNNs) have become the dominant paradigm in artificial intelligence, achieving remarkable performance across computer vision, natural language processing, and reinforcement learning. However, their primary learning algorithm - error backpropagation \citep{rumelhart1986learning, werbos1994roots} - presents significant challenges for practical deployment and biological plausibility \citep{zarkeshian2022photons, lillicrap2020backpropagation}. Backpropagation is a supervised learning process that relies on propagating error gradients backward through the network’s layers to adjust connection weights so that the network outputs come close to some given target output. This global update process requires storing all intermediate activations and computing gradients across the entire network \citep{rumelhart1986learning}.

These constraints impose several critical limitations: the algorithm's requirement for precise error signals demands massive labelled datasets \citep{lagani2022recent}, while the global nature of weight updates leads to extreme computational costs and energy consumption \citep{wang2024ecological, strubell2020energy}. Additionally, the sequential backward pass creates bottlenecks resulting in lengthy training times \citep{krithivasan2022accelerating}, and the rigid supervised learning approach limits adaptability while leaving models vulnerable to adversarial attacks     \citep{lee2018simple}.

Conversely, Hebbian learning, a biologically inspired learning algorithm, circumvents the need for error backpropagation, updating weights locally, using unsupervised local neural activity correlations. Although Hebbian learning shows promise in addressing these challenges, posing as an alternative that mirrors the efficiency and adaptability of biological systems \citep{gerstner2002mathematical}, levels of accuracy as those reported with backpropation are still out of reach \citep{demidovskij2023implementation}.

Recent research in Hebbian learning has focused on combining purely Hebbian learning for unsupervised feature extraction in convolutional layers, with  supervised classification training on these frozen features. Successful Hebbian integration into DNNs \citep{amato2019hebbian, krotov2019unsupervised, miconi2021hebbian, journe2022hebbian} is primarily achieved through neural competition, which allows distinct neurons to respond to input patterns, preventing redundant feature learning and promoting the development of diverse hierarchical representations. Two representation coding schemes have been used: soft competition methods \citep{journe2022hebbian} that enable distributed updates across all neurons, as well as several hard competition approaches \citep{grinberg2019local, amato2019hebbian, miconi2017biologically} where neurons undergo sparse updates that limit the number of participating neurons.

Whereas soft-Winner-Takes-All (WTA) competition through  \cite{journe2022hebbian}'s SoftHebb model achieves an accuracy on CIFAR-10 of nearly 80\%, current hard-WTA Hebbian approaches face challenges to match backpropagation SOTA performance while maintaining biological plausibility and efficiency. The SOTA accuracy with a Hard competition WTA model (Hard-WTA) on CIFAR-10 is 72.2\% \citep{grinberg2019local}. Although this architecture utilised just one convolutional layer, it does not facilitate the construction of hierarchical representations of the input, which are crucial for developing intricate representations and understanding the relationships within the data. Multilayer networks are necessary to achieve this. A sparse 3-CNN layer Hebbian architecture led to only 64.6\% accuracy \citep{miconi2021hebbian}, and struggled to form meaningful hierarchical representations. 

Arguably, sparse neural coding derived from Hard-WTA has greater biological plausibility \citep{barth2012experimental} paralleling the highly sparse brain connectivity with less than 1\% neurons firing in the visual cortex when representing an image \citep{yoshida2020natural}.
This sparseness facilitates the energy efficiency and high speeds of biological systems. However, local sparse representations in ANNs generally suffer from poor generalisation and scaling, resulting in decreased performance with each additional layer in the network \citep{lagani2022comparing}.

Our research addresses these Hard-WTA challenges by advancing the integration of Hebbian learning into modern deeper CNN architectures for unsupervised feature extraction through three key contributions. First, we integrated and optimised the computational efficiency of Hard-WTA in \cite{journe2022hebbian}'s architecture. Second, we furthered biological realism through Bienenstock, Cooper, and Munro (BCM) learning rule \citep{bienenstock1982theory}, novel spatial lateral inhibition, pre-synaptic, temporal, and homeostatic competition mechanisms that mirror cortical processing \citep{blakemore1976synaptic}.Third, we developed architectural improvements including depthwise separable convolutions and residual connections, reducing parameter count by 6.6x.

To assess performance and explore a suitable Hard-WTA model and architecture integrating these individual innovations, we implemented 14 distinct Hebbian feature extraction configurations, and a comparative end-to-end backpropagation model. These configurations used a two-phase training approach: unsupervised Hebbian learning for feature extraction in convolutional layers, followed by a supervised classifier layer trained via backpropagation. This approach was chosen to match earlier Hebbian CNN research models, which while maintaining biological plausibility in the feature learning phase, enable direct comparisons with conventional end-to-end backpropagation CNNs.

By integrating our best-performing learning mechanism (lateral inhibition and BCM Hebbian learning) with Hard-WTA, we achieved 75.2\% CIFAR-10 test mean accuracy on the final 50\% epochs in \cite{journe2022hebbian}'s 3-CNN layer architecture, setting a new SOTA Hard-WTA performance. This represents a 10.6\% improvement over the previous Hard-WTA SOTA Hebbian performance of 64.6\% using the same network depth \citep{miconi2021hebbian}

Our experimental results evidenced that Hebbian learning for feature extraction leads to a comparable performance to end-to-end backpropagation methods, which achieved 75.2\% mean accuracy under the same training conditions (20 epochs and same architecture). Our advancements were also sustained when additional datasets were used, with the Optimal-HardWTA setup reaching 98\% test mean accuracy on MNIST and 69.5\% on STL-10.

These results significantly advance the SOTA for Hard-WTA Hebbian models, addressing key challenges faced by the Hebbian-AI community: maintaining competitive performance across multiple layers while reducing computational requirements and enhancing biological realism.

 To validate our implementation and results, we developed a comprehensive evaluation framework that incorporates a PyTorch-based, modular implementation enabling direct integration and comparison of different Hebbian approaches alongside visualisation tools such as Uniform Manifold Approximation and Projection (UMAP \citep{mcinnes2018umap}) for feature embeddings, receptive fields and weight distributions of neurons for the analysis of competition and feature learning.

\section{Background and Related Work}\label{context}

\subsection{Convolutional Neural Networks}

Neurons, both biological and artificial, function as fundamental units of information processing. In Artificial Neural Networks (ANNs), neurons are mathematical models mimicking the behaviour of their biological counterparts. Learning in both systems involves modifying connection strengths or synapses between neurons. 

 In an ANN, a neuron pass operates only on local information during a forward pass, taking, taking an input vector $x$, applying synaptic weights $w$, and producing an output $y = f(\sum x^T w)$, where $f$ is an activation function. This output is then compared against a target outcome (ground truth) and the error is used to update weights globally through the network via the backpropagation algorithm and gradient descent optimisation rule. \citep{rumelhart1986learning, rojas1996backpropagation, amari1993backpropagation}.

Convolutional Neural Networks (CNNs) \citep{lecun1998gradient} are deep ANN architectures built on key principles of local receptive fields and hierarchical feature extraction. These architectural elements, inspired by the organisation of the cat's visual cortex \citep{fukushima1980neocognitron}, make CNNs particularly effective for visual processing tasks. In CNNs, neurons are structured into layers for hierarchical processing of visual data. At every level, neurons generate responses to patterns in their receptive field by aggregating inputs from units in a preceding layer, resulting in an expanded receptive field and learning progressively complex and abstract hierarchical relationships or representations from the data, with lower layers typically extracting simple patterns like edges or textures while higher layers combine these into sophisticated representations such as shapes or objects, thus enabling more effective understanding and classification.

While CNNs have traditionally been trained using backpropagation, their fundamental architectural principles are valuable regardless of the learning algorithm
employed. Modern CNN models often incorporate additional features, such as integrated residual blocks and Depthwise Separable Convolutions (DSC), inspired by biological processes to enhance their effectiveness.

Residual blocks allow training for very deep backpropagation networks by providing skip connections between layers \citep{he2016deep} mimicking the hierarchical feedback pathways found in biological visual systems. DSCs separate a typical convolution operation into two separate convolutional operations \citep{chollet2017xception} for computational efficiency and preventing overfitting through parameter reduction.
These features can also be used alongside Hebbian learning. Hence, following the rationale to improve biological plausibility and efficiency, we integrated residual blocks and DSC into our Hebbian-CNN framework.

Our depthwise implementation follows \cite{chollet2017xception} Depthwise Separable Convolution (DSC) approach, which modifies standard CNN operations by dividing them into two steps: depthwise convolution (applies a single filter per input channel) and pointwise convolution (applies 1x1 convolution across all channels). This approach not only reduces parameters and computations while maintaining performance but also increases biological plausibility \citep{tomen2021deep}. DSC achieves this through the separation of spatial and feature combination operations, sparser connections between layers, and independent processing of each channel. Recent research \citep{babaiee2024neural} has shown that kernels in backpropagation-trained depthwise separable networks exhibit centre-surround receptive fields similar to those found in biological visual systems.

\subsection{Fundamental Challenges and Biological Implausibilities of Backpropagation}

Backpropagation faces fundamental optimisation challenges that limit its effectiveness. The algorithm often struggles with local convergence in nonlinear optimisation problems, frequently becoming trapped in suboptimal local minima \citep{sexton2000comparative}. Recent critiques have highlighted additional fundamental issues: backpropagation requires unrealistic precision in weight updates, cannot effectively handle temporal dependencies, and struggles with credit assignment across multiple timescales \citep{hinton2022forward}. Despite its effectiveness in training ANNs, backpropagation is assumed to posses little biological plausibility. \citep{apparaju2022towards, lillicrap2020backpropagation, song2020can, NEURIPS2023_12033923}.

Backpropagation requires error signals from non-directly connected neurons \citep{lillicrap2020backpropagation}, whereas biological systems rely on local interactions for synaptic plasticity. It needs the transpose of the weight matrix during backward pass, while biological neurons use unidirectional synapses (Weight Transport problem) \citep{apparaju2022towards}. The error computation requires complete forward and backward passes, unlike biological neurons, which do not exhibit such precise coordination (Update Locking problem) \citep{song2020can}. It relies on top-down supervision signal through a global loss function \citep{lillicrap2020backpropagation}, whereas biological learning involves more localised error correction.

Dale's Principle \citep{eccles1976electrical}, a fundamental concept in neuroscience, states that neurons release the same neurotransmitters at all their synapses, leading to either exclusively excitatory or inhibitory effects. This biological constraint is typically violated in artificial neural networks trained through backpropagation \citep{cornford2020learning}, where single neurons can have both positive and negative weights.

These biological limitations directly translate to implementation challenges in neuromorphic computing \citep{schuman2022opportunities}, creating bottlenecks in parallel processing, while its heavy energy and memory requirements, and synchronised update needs limit deployment on edge devices and real-time applications.

\subsection{Hebbian Learning}

Backpropagation's biological limitations substantiate the significance and potential of Hebbian learning for theoretical research and practical applications. Hebbian learning \citep{hebb2005organization, lagani2023synaptic}, follows the principle \textit{neurons that fire together, wire together} \footnote{Originally formulated as "When an axon of cell A is near enough to excite a cell B and repeatedly or persistently takes part in firing it, some growth process or metabolic change takes place in one or both cells such that A’s efficiency, as one of the cells firing B, is increased. \citep[p. 62]{hebb2005organization}" }. Neural updates are reliant solely on the activity of locally connected neurons. These neurons undergo a completely unsupervised learning process, as they do not require input from an external teacher or an error signal, and can identify statistical patterns in input data. The local nature of Hebbian learning eliminates the need for backward passes and target signals through the network, reducing memory requirements and enabling parallel weight updates. This locality makes Hebbian learning particularly attractive for neuromorphic hardware implementation \citep{schuman2022opportunities}, edge computing applications, and scenarios with limited labelled data.

The biological soundness of Hebbian learning, specifically its continuous adaptation to input patterns and local learning rules, enables systems to naturally adjust to changing input distributions and maintain stability without global coordination. These properties contrast with backpropagation systems which typically require complete retraining to adapt to new patterns and can become unstable when input distributions shift from their training data.

The development of Hebbian learning rules has evolved to address key challenges in neural network stability and functionality. The basic Hebbian rule, which forms the foundation of this evolution, is formally expressed as:

\begin{equation}\label{basic}
    \Delta w(t) = \eta y(x,w) x
\end{equation}

where $y(x, w)$ or $y$ is the post-synaptic activation or output of the neuron, $x$ is the pre-synaptic activation or input of a neuron, and $\eta$ the learning rate. While this basic rule captures the essence of Hebbian learning by strengthening connections between co-active neurons, it suffers from unbounded weight growth or Long Term Potentiation (LTP), leading to instability. This limitation prompted the development of three significant variants that we examine here for their relevance to our work in stable learning dynamics.

Due to its simplicity and effectiveness in deep network architectures, we build our implementation upon Grossberg's Instar rule \citep{grossberg1976adaptive}. This rule introduces weight $w$ as a decay proportional to post-synaptic activity $y$, enabling both weight growth and decay based on input-output correlations:

\begin{equation} \label{grossberg}
\Delta w(t) = \eta y (x-w) = \eta (yx- yw)
\end{equation}

The equation is presented in two mathematically equivalent forms to emphasise different interpretations of the learning process. The first expression, $\eta y (x-w)$, highlights the biological intuition: the weight change is proportional to the difference between the input signal $x$ and the current weight $w$, gated by the post-synaptic activity $y$. The second expression, $\eta (yx- yw)$, reveals the rule's dual mechanism: a Hebbian term $yx$ for weight growth and an normalisation term $yw$ for weight decay or Long Term Depression (LTD), providing inherent stability without requiring complex normalisation schemes or additional parameters.

Our model also implements the BCM rule \citep{bienenstock1982theory} due to its strong biological plausibility and effectiveness in preventing runaway synaptic growth. The rule's dynamic threshold mechanism closely mirrors biological synaptic modification processes, making it particularly suitable for our focus on biologically-inspired learning systems. Belonging to the family of pre-synaptic gating rules, it inverts the roles of $x$ and $y$ from Equation \ref{grossberg}:

\begin{equation}\label{bcm}
\Delta w(t) = \eta x \psi(y-\theta) = \eta x y(y-\theta)
\end{equation}

The equation is presented in two equivalent forms to highlight different aspects of the rule. The first expression uses the nonlinear function $\psi$, which represents the general form of the modification threshold function. In the specific implementation we adopt, this function takes the form $\psi(y-\theta) = y(y-\theta)$, leading to the second expression. This quadratic form ensures that weight modifications exhibit both potentiation and depression depending on the post-synaptic activity level relative to the threshold $\theta$, creating a natural stability mechanism in the learning process.

While Sanger's rule \citep{sanger1989optimal}, an extension of Oja's original work \citep{oja1982simplified}, achieves weight normalisation and enables online Principal Component Analysis, its requirement for non-local synaptic information $\sum y_j w_j$ makes it less suitable for biologically plausible learning mechanisms and thus is not implemented in our model. For completeness, the rule is expressed as:

\begin{equation}
\Delta w_i(t) = \eta y_i(x-\sum_{j=1}^i y_j w_j)
\end{equation}

This theoretical Hebbian foundation underpins the weight update rule in Hopfield networks \citep{hopfield1982neural}, where the Hebbian principle is applied to store patterns as attractors in the network. Variants like dense associative memory \citep{krotov2023new} extend this connection by leveraging nonlinear interactions, further demonstrating the versatility of Hebbian principles.

\subsection{Neural Competition and Stability Mechanisms}

Neural competition ensures different neurons learn to respond to different input patterns, preventing redundant feature learning. Through competitive mechanisms, neurons in each layer specialise in detecting distinct features, from simple edges and textures in early layers to more complex patterns in deeper layers. This specialisation enables the emergence of hierarchical representations, where higher layers combine and build upon the features detected by lower layers. While Hebbian learning provides biologically-plausible learning, it tends toward instability and redundancy without competition, as neurons typically converge to respond to the same dominant features in the input. We  implemented and compared several competition and stability mechanisms to investigate their performance and adequacy. 

Winner-Takes-All (WTA) competition serves as the primary mechanism for driving neural competition in Hebbian learning, with two main variants: A) Hard WTA \citep{rumelhart1985feature} allows only the neuron with the maximum activation (winning neuron) to update weights, promoting sparse representations and distinct feature specialisation. This mechanism can be applied within layers, across channels, or the whole network, providing flexibility in how competition is structured. B) Soft WTA \citep{nowlan1989maximum} uses a softmax function with a temperature parameter to allow distributed learning while maintaining competition, providing more nuanced feature representations at the cost of more complex parameter tuning.

We investigated the integration of other competition and stability strategies with WTA competition architectures. New architectural configurations were thus built by independently combining: C) Anti-Hebbian learning \citep{choe2022anti, foldiak1990forming} which actively decorrelates neural activities by weakening synaptic strengths between co-activated neurons with lateral connections, supporting the WTA-driven specialisation through explicit decorrelation D) Lateral inhibition \citep{gabbott1986quantitative} enhances WTA competition by modulating neuron activity based on neighbouring neurons' activity levels, particularly valuable in visual processing tasks where local contrast enhancement directly influences feature detection quality. E) Pre-synaptic competition \citep{rasmussen1993presynaptic} reflects biological systems' resource constraints and complements WTA by regulating input signal competition, affecting learning outcomes and representation efficiency.

Our network stability enhancements focused on two independent key mechanisms which neural systems employ to regulate activity: synaptic traces and homeostatic plasticity. F) We implemented synaptic traces \citep{morris2006elements}, a form of short-term plasticity,which enable neurons to retain information about their recent activation history, influencing their response to subsequent inputs. This temporal integration of activity plays a crucial role in learning and memory formation. G) We also employed homeostatic plasticity, \citep{turrigiano2004homeostatic} which represents another fundamental biological mechanism where neurons adjust their properties to maintain stable activity levels while preserving their ability to respond to relevant stimuli. This process helps prevent neural overexcitation and ensures efficient coding of input patterns across varying conditions.

\subsection{Previous Hebbian-CNN Integration}

Recent research has incorporated Hebbian learning principles and neural competition into CNNs, with varying approaches to Winner-Take-All (WTA) competition yielding different trade-offs between performance and biological plausibility. A key characteristic in these approaches is the separation between unsupervised feature learning and supervised classification. The convolutional layers learn representations through purely local Hebbian updates without labels or error signals, while only the final classifier layer uses supervised learning. For proper comparison with existing research, our models followed the same general scheme.

\cite{journe2022hebbian} set the current Hebbian benchmark with 80\% accuracy on CIFAR-10. Their implementation combined soft-WTA competition with subtractive weight competition learning, which is mathematically akin to a variant of the Grossberg Instar rule.

\begin{equation}\label{softhebb_eq}
\Delta w_{ik}^{(SoftHebb)} = \eta \cdot y_k \cdot (x_i - u_k \cdot w_{ik}).
\end{equation}

where $w_{ik}$ is a synaptic weight from a pre-synaptic neuron $i$ with activation $x_i$, $u_k$ is the original post-synaptic output of neuron $k$, and $y_k$ is the result of the post-synaptic softmax competition. While this leading approach represents significant progress, it required a fourfold increase in the neurons per layer, highlighting ongoing efficiency challenges.

\cite{grinberg2019local} achieved the leading performance in Hard-WTA, implementing a single CNN layer architecture with patch normalisation, reaching 72.2\% accuracy on CIFAR-10 (a natural image dataset). \cite{amato2019hebbian} further advanced the hard-WTA approach with cosine similarity activation, achieving 98.55\% accuracy on MNIST (a simple dataset of handwritten digits) but only 64\% on CIFAR-10 using a two-layer CNN. Data whitening emerged as an essential pre-processing step for enhanced performance with hard-WTA competition. Their work revealed fundamental limitations in scaling hard-WTA to deeper architectures, as its performance markedly decreased with
additional layers. Building on this foundation, \cite{lagani2022comparing} maintained 60\% accuracy on CIFAR-10 with a deeper five-layer model by implementing Sanger's rule.

\cite{miconi2021hebbian} reached the highest Hard-WTA accuracy for a shallow network with 64.6\% in a three CNN-layer model using a hybrid approach combining hard-WTA, homeostasis, new Triangle Activation function, and extensive pruning. Despite this improvement in accuracy in a deeper architecture, it is trained through gradient-based backpropagation learning, using surrogate losses which are equivalent to Hebbian learning rules.

Still, these Hebbian approaches incorporating WTA competition demonstrate Hebbian learning's advantages in convergence speed, performance with limited data, and resilience against adversarial attacks \citep{gupta2022bio}. Recent research \citep{laganireserach} suggests several promising directions for improvement, including incorporating inter-layer feedback and top-down connections, implementing pre-synaptic competition, applying the BCM learning rule, and developing spatial decorrelation in WTA competition.

Our work addressed the challenge of scaling hard-WTA competition in deeper architectures, leveraging its key advantages over soft-WTA: it reflected the binary firing states of biological neurons, enables energy-efficient implementations, and crucially, enforces true sparseness by silencing all but the strongest neurons.

Integration of the theoretical foundations of Hebbian learning, neural competition, and stability mechanisms from the biological neural networks into modern convolutional neural networks promises more efficient and biologically plausible deep learning systems, allowing networks to learn from limited data, exhibit greater noise robustness, and consume fewer computational resources. This approach bridges biological and artificial systems, potentially enabling more resilient and flexible learning. In this paper, we first searched for a suitable combination of these mechanisms, exploring their accuracy and potential to enhance the formation of hierarchical representations, incorporating hierarchical layers of increasingly abstract, sparse representations shaped by competition. Then we selected the 4 most suited configurations based on their accuracy for further evaluation and analysis.

\section{Methodology}\label{methodology}

This section outlines the PyTorch-based Hebbian-CNN framework adopted. We designed and implemented 15 distinct configurations to systematically evaluate their components and interactions and build a naturalistic optimal architecture based on their performance as measured by accuracy values. To this aim, we developed a configurable architecture using different learning algorithms, Hebbian Grossberg Instar and BCM learning, two main neural competition mechanisms, hard and soft WTA, and additional temporal, homeostatic, pre-synaptic and spatial competition mechanisms. These mechanisms were implemented as optional components in a customisable Hebbian layer. Each element could be enabled, disabled, or combined into distinct configurations to allow us to explore their performance. Architectural changes, including depthwise separable convolutions and its enhancement through skip connections in residual blocks were evaluated.

The design philosophy behind this work encompasses computational efficiency, code readability, modular architecture, parametric flexibility, and extensibility. We prioritised a GPU-accelerated implementation leveraging the PyTorch framework, known for its efficient parallel processing capabilities in deep learning applications. The modular structure closely mirrors that of PyTorch, providing an intuitive organisational system familiar to researchers experienced with this machine learning library. The code is available on : \href{https://github.com/Julian-JN/Advancing-the-Biological-Plausibility-and-Efficacy-of-Hebbian-Convolutional-Neural-Networks}{GitHub}

\subsection{Learning Pipeline}

A two-phase approach was used to ensure feature learning  biological plausibility while having a standard classifier system. Unsupervised Hebbian learning was employed for feature extraction in convolutional layers, followed by supervised backpropagation linear classifier training on the frozen features.

\begin{figure}[h!]
  \centering
  \resizebox{0.99\textwidth}{!}{%
  \begin{minipage}[b]{0.99\textwidth}
      \includegraphics[width=\textwidth]{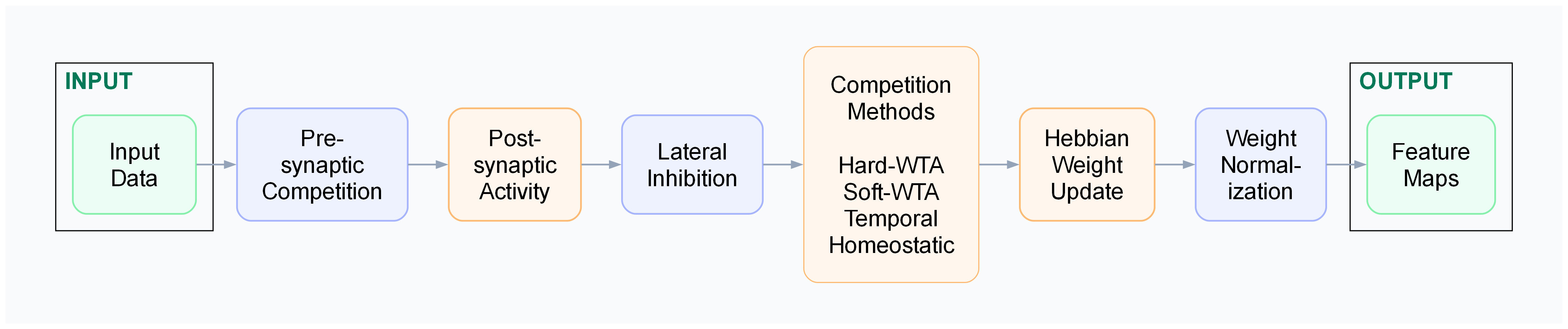}
      \centering
    \end{minipage}
  }
  \caption{Diagram of the Hebbian learning process in our custom Hebbian-CNN layer. Green boxes correspond to input/output, blue boxes indicate optional processing, and orange boxes designate core learning and competition mechanisms.}
  \label{fig:learning-diagram}
\end{figure}

The framework implemented a consistent, unsupervised feature learning pipeline across all Hebbian configurations, enabling feature extraction without requiring labelled data or global error signals. Figure \ref{fig:learning-diagram} illustrates the Hebbian learning process in each CNN layer. Each forward pass followed these steps:

\begin{enumerate}
    \item Pre-synaptic competition is applied to modify weights $w$ before any calculations if enabled
    \item Post-synaptic activities are computed through convolution between modified weights and input
    \item Lateral inhibition modulates post-synaptic activities through the DoG kernel when enabled
    \item Post-synaptic competition mechanisms are applied when enabled:
    \begin{itemize}
        \item Hard-WTA selects maximally activated neurons
        \item Temporal Competition uses synaptic traces for temporal specialisation
        \item Homeostatic Competition adjusts selectivity based on input statistics
    \end{itemize}
    \item Weights are updated using the chosen learning rule (Grossberg, BCM, or SoftWTA Grossberg)
    \item Weight normalisation is applied to maintain stable learning dynamics
\end{enumerate}

This pipeline can be configured through layer parameters to activate different combinations of mechanisms based on research requirements.

\subsection{Dataset Selection and Pre-Processing}

To evaluate our framework's performance and compare it with existing Hebbian CNN approaches, CIFAR-10 \citep{krizhevsky2009learning}, MNIST \citep{mnist} and STL10 \citep{coates2011analysis} were chosen as the benchmark dataset. The initial experimental configurations aimed at selecting the optional learning mechanisms and replicating the literature conducted on CIFAR-10. CIFAR-10 consists of 60,000 RGB images of 32x32 pixels, MNIST consists of 70,000 grayscale images of 28x28 pixels, and STL-10 consists of 13,000 RGB images of 96x96 pixels across 10 classes. Data processing included implementing random horizontal flip augmentations and zero-phase component analysis (ZCA) whitening. ZCA whitening was computed as improvements in performance were found for Hard-WTA competition \citep{amato2019hebbian}.

\subsection{Core Learning and Competition Mechanisms}

Previous non-gradient-based implementations of synaptic plasticity using the Grossberg Instar rule with Hard-WTA \citep{amato2019hebbian} and Soft-WTA \citep{journe2022hebbian} were built into the system to be replicated as experimental controls and foundational elements for investigating enhanced biological learning approaches.

\subsubsection{Grossberg Hebbian Learning Implementation}

The Grossberg Instar stability modification of the basic Hebbian rule was used as the primary learning algorithm. The process began with computing post-synaptic activities through convolution:
\begin{equation}
y = w * x
\end{equation}

where $*$ denotes the convolution operation between weights $w$ and pre-synaptic activities $x$.

Following Lagani's approach \citep{lagani2022comparing}, we refined Hard-WTA neural selectivity by computing the post-synaptic activity using a cosine similarity function involving the postsynaptic convolution operation. With this approach, neurons fire strongly only when inputs closely match the learnt feature, increasing feature specialisation. The post-synaptic activity $y$ is calculated as:

\begin{equation}
y = \text{sim}(x, w) = \frac{x * \hat{w}}{\|x\|},
\end{equation}
where $x$ is the input tensor, \( \hat{w} \) is the L2-normalised weight kernel: 
\[\hat{w} = \frac{w}{\|w\|_2}\]

The operation \( x * \hat{w} \) denotes the 2D convolution, which naturally handles the different shapes of input and weight tensors by sliding the weight kernel across the input.

The norm \( \|x\| \) is computed locally for each position where the convolution is applied:
\[\|x\| = \sqrt{x^2 * \mathbf{1} + \epsilon}\] 
where \( \mathbf{1} \) is an all-ones filter with the same shape as $w$, and \( \epsilon = 10^{-8} \) prevents division by zero. This ensures that the computed similarity aligns with classical cosine similarity but is applied locally in a convolutional context.

The Grossberg weight update rule was then calculated for the convolutional neural network as:

\begin{equation}
\Delta w = (y * x) - (\sum y) w. \label{grossberg2}
\end{equation}

The derivation of Equation \ref{grossberg2} from Equation \ref{grossberg} can be found in Appendix \ref{derivation} . These updates were normalised and stored in a buffer for efficiency and analysis.

\subsubsection{Hard-WTA Competition}
Hard Winner-Takes-All (WTA) competition was implemented to achieve sparse and specialised feature representations. For each spatial location $(h,w)$, competition took place across channels to create a binary mask $\text{mask}_{WTA}(y)$ that outputs 1 for the maximum activation across channels and 0 otherwise. The final post-synaptic activities are then computed as:

\begin{equation}
y_{final} = y \odot \text{mask}_{WTA}(y)
\end{equation}

Our implementation enhanced Amato's approach \citep{amato2019hebbian} by harnessing PyTorch's GPU matrix and convolutional operations for efficient parallel computation, as suggested by \cite{lagani2022fasthebb}. This allowed weights to update during the forward pass without requiring gradients or backpropagation, significantly improving computational efficiency while maintaining biological plausibility.

\begin{figure}[h!]
\centering
\resizebox{0.8\textwidth}{!}{%
\begin{minipage}[b]{0.8\textwidth}
\includegraphics[width=\textwidth]{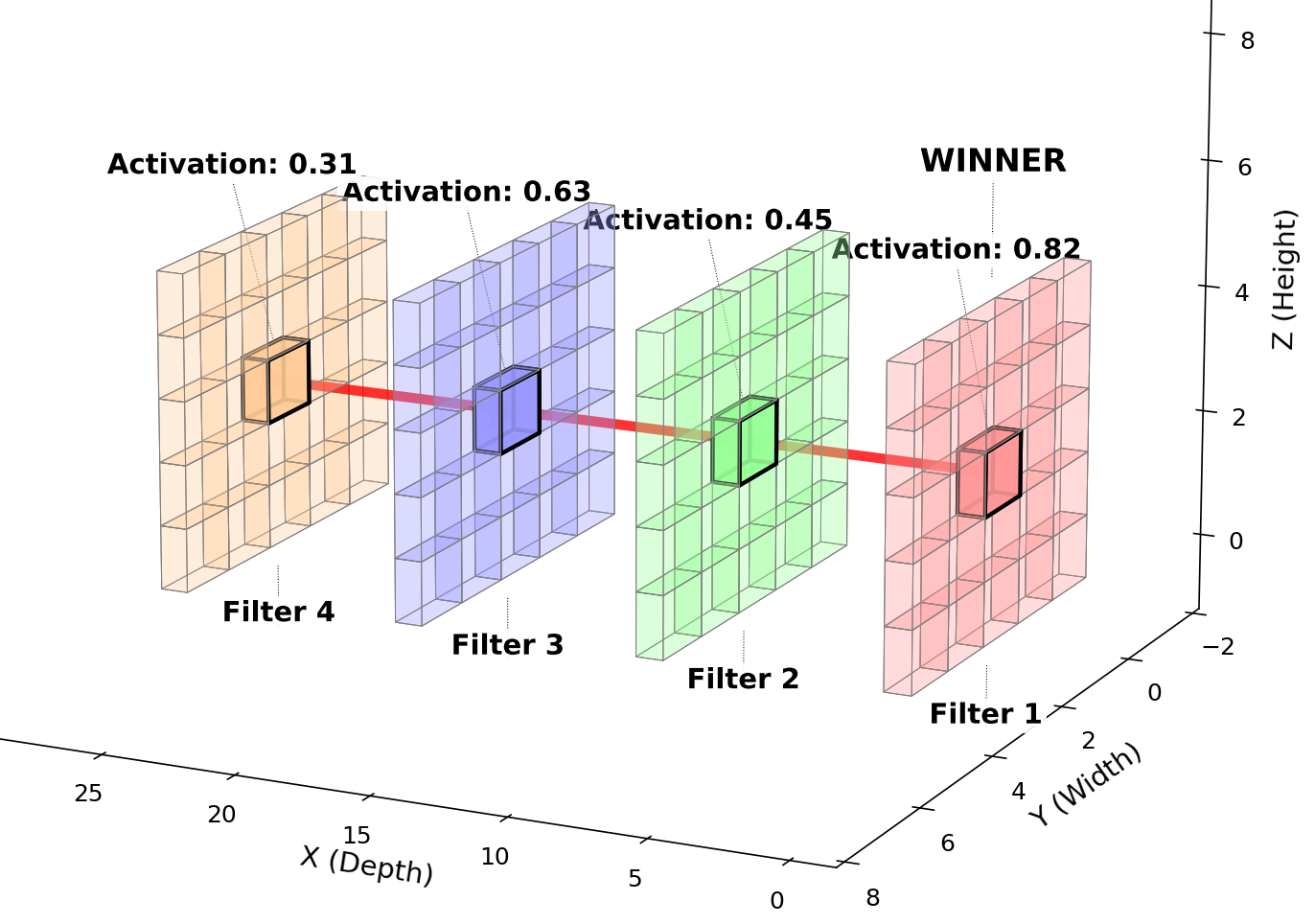}
\centering
\textbf{(A) Hard-WTA Competition}
\end{minipage}
}
\caption{Implementation of Hard-WTA competition in CNNs influences neurons situated at the same spatial position across various filters.}
\label{fig:wta-comp}
\end{figure}

\subsubsection{SoftHebb Implementation}

To establish a comprehensive comparison with current state-of-the-art Hebbian methods, we implemented SoftHebb as detailed in \cite{journe2022hebbian}. Unlike in Hard-WTA, SoftHebb applied soft competition to postsynaptic activity through a Softmax function with an inverse
temperature parameter (Equation \ref{softhebb_eq}). A key distinction of this method is its use of a simple form of anti-Hebbian learning (negating SoftHebb’s weight update) for all neurons except the maximally activated one, creating a more nuanced form of competition than the binary selection of Hard-WTA (p.4 \cite{journe2022hebbian}).

The combination of these three core procedures (learning algorithm and hard/soft-WTA) provided the foundation for our framework and the modularity of its components allowed for a systematic evaluation of their individual and combined effects on the network's performance.

\subsection{BCM Learning Rule}

We implemented the BCM learning rule (Equation \ref{bcm}) within CNNs to evaluate its unsupervised feature learning capacity compared to standard Hebbian approaches. Our convolutional implementation started by computing winner-take-all (WTA) activities for each neuron:
\begin{equation}
y_{WTA} = y \odot \text{mask}_{WTA}(y)
\end{equation}
where $\odot$ represents element-wise multiplication. For each output channel, we maintained an adaptive threshold $\theta$ that tracked the average squared activity of WTA neurons through an exponential moving average:
\begin{equation}
\theta_t = (1 - \alpha)\theta_{t-1} + \alpha \mathbb{E}[y_{WTA}^2]
\end{equation}
where $\alpha$ is the \path{theta_decay} parameter controlling adaptation speed (set at a default value of 0.5), and $\mathbb{E}[\cdot]$ denotes the spatial mean. The BCM non-linearity $\psi(y-\theta)$ was then computed as:
\begin{equation}
\psi(y-\theta) = y_{WTA} \odot (y_{WTA} - \theta)
\end{equation}

The final weight updates $x\psi(y-\theta)$ were computed by convolving the presynaptic input $x$ with the BCM non-linearity and normalising to maintain consistent weight magnitudes across layers. Our method preserved the Hard-WTA competition mechanism, substituting the Grossberg rule, commonly employed in traditional Hebbian learning, with BCM's adaptable threshold dynamics.

\subsection{Temporal and Homeostatic Competition}


Two competition strategies, grounded in biological stability principles and relying on the competitive model proposed by \cite{laganireserach}, were implemented among neurons situated in the same spatial area but across channels to enhance biological realism and robustness.

We developed two selection mechanisms following Lagani's proposals. Our \textit{Temporal Competition} implementation tracked neuronal activity through an activation history buffer $H_t$ of size $T$ for each neuron:
\begin{equation}
H_t = {y_{t-T}, y_{t-T+1}, ..., y_t}
\end{equation}
From this buffer (with default size of 500), we computed a temporal threshold $\theta_t$ as the median of historical activations:
\begin{equation}
\theta_t = \text{median}(H_t)
\end{equation}
This threshold-based approach promoted neurons demonstrating consistent activity over time, as only neurons whose activation $y$ exceeded $\theta_t$ are permitted to update their weights.

For \textit{Homeostatic Competition}, we applied an input-driven competition mechanism that calculates normalised similarity scores between weights and input:
\begin{equation}
S_{c,h,w} = \frac{\langle x, w \rangle}{|w|_2 + \epsilon}
\end{equation}
where $S_{c,h,w}$ represents these scores across channels and spatial locations, with $\epsilon = 10^{-10}$. Based on these similarities, we computed an adaptive threshold:
\begin{equation}
\theta_{adaptive} = \mu_S + k\sigma_S
\end{equation}
where $\mu_S$ and $\sigma_S$ are the mean and standard deviation of similarities respectively, and $k$ is the competition factor (typically set to 2). This adaptive threshold dynamically adjusted neural plasticity, increasing when input patterns produced high similarity scores to prevent overactivation, and decreasing with low scores to facilitate more neural updates. Only neurons with post-synaptic activities $y$ exceeding $\theta_{adaptive}$ updated their weights.

Both selection mechanisms were designed to be compatible with additional competition methods like Hard-WTA to further promote efficient specialisation and sparse connectivity. We note these competitive selection mechanisms may limit BCM's inherent anti-Hebbian learning for sub-threshold neurons, potentially affecting feature specialisation that would normally occur through negative weight updates in traditional BCM implementations.

\subsection{Pre-synaptic Competition}
Pre-synaptic competition was enforced using synaptic couplings, modifying the weights $w$ before calculating the postsynaptic activity. Three competition modes were developed across the input channel, where different input features compete to influence the same output. Each mode first computed an inverse weight magnitude ($m = \frac{1}{|w| + \epsilon}$) to promote competition between input features with weaker connections, drawing inspiration from homeostatic synaptic plasticity mechanisms described by \cite{turrigiano2012homeostatic}. This inverse weighting allows less established pathways to maintain competitive influence, though it represents a conflict with Hard-WTA approaches that exclusively reinforce dominant connections. Here, $\epsilon$ is a small constant (1e-6) to prevent division by zero:

\begin{enumerate}
\item \textbf{Linear Competition, defined as:}
\begin{equation}
w_{eff} = \frac{m}{\sum_{i} m_i + \epsilon}
\end{equation}
It normalises the weights to values between 0-1.
\item \textbf{Softmax Competition, defined as:}
\begin{equation}
w_{eff} = \text{softmax}(m) = \frac{e^{m_i}}{\sum_{j} e^{m_j}}
\end{equation}
This creates a more pronounced competition where stronger connections are emphasised, and the weights summed together add up to 1.
\item \textbf{L2 Norm Competition, defined as:}
\begin{equation}
w_{eff} = \frac{m}{\sqrt{\sum_{i} m_i^2}}
\end{equation}
This ensures that the sum of squared effective weights equals 1.
\end{enumerate}

\subsection{Lateral Inhibition}

A fixed kernel created through a difference of Gaussians (DoG) function was designed to simulate lateral inhibition and surround modulation, as proposed by \citep{hasani2019surround}. This kernel models how neurons in the visual cortex respond to stimuli in their receptive field and surrounding areas. The centre (excitatory) region represents the classical receptive field, while the surround (inhibitory) region models lateral inhibition from neighbouring neurons. Mathematically it is defined as :

\begin{equation}\label{lateral_kernel}
    K_{SM}(x,y) = \frac{1}{K_{center}} \left( \frac{G_e(x,y)}{2\pi\sigma_e^2} - \frac{G_i(x,y)}{2\pi\sigma_i^2} \right)
\end{equation}

where $G_e$ and $G_i$ represent the excitatory and inhibitory Gaussian functions, respectively, and $\sigma_e$ and $\sigma_i$ denote the standard deviations of the excitatory and inhibitory Gaussians (with default value of 1.2 and 1.4). This kernel strengthens synapses in the immediate spatial neighbourhood of a neuron, and weakens synapses further from this neighbourhood. When applied, lateral inhibition was computed by applying a convolutional kernel of size 5, with filter in the form of Equation \ref{lateral_kernel}.

\subsection{Architectures}

\begin{figure}[h!]
  \centering
  \resizebox{0.999\textwidth}{!}{%
  \begin{minipage}[b]{0.99\textwidth}
      \includegraphics[width=\textwidth]{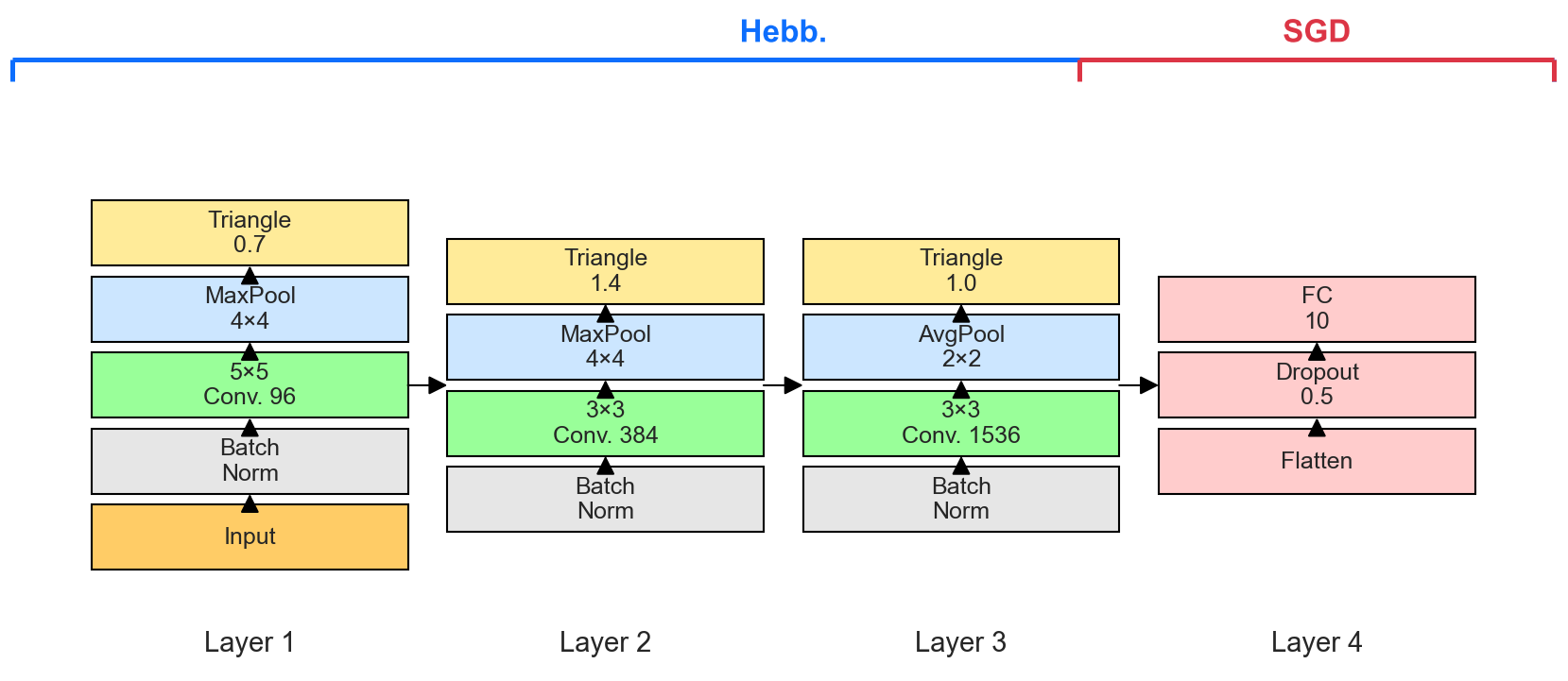}
      \centering
      \textbf{Journe Architecture}
    \end{minipage}
  }
  \caption{Visual representation of the main Journe Architecture, composed of 3 Hebbian CNN layers and a final backpropagation linear classifier layer.}
  \label{fig:architecture}
\end{figure}

Two main architectures were implemented, Journe's and Lagani's. The former was the default architecture used in all configurations unless stated otherwise. To investigate biologically-plausible learning mechanisms while maintaining computational efficiency, we explored three distinct architectural variants, specifically Journe along with its counterparts, the depthwise separable convolution, and the residual block. Each variant built upon Journe baseline architecture while introducing specific modifications aimed at reducing parameters, improving performance, and enhancing biological realism. These architectures were evaluated independently to assess their individual contributions to the model's capabilities.


The baseline architecture, Journe (Figure \ref{fig:architecture}), consisted of 3 convolutional layers featuring padding, Triangle/Rectified Polynomial Unit activation function \citep{miconi2021hebbian, journe2022hebbian}, and an increased number of filters in each subsequent layer (quadrupling the channels at each layer). This design choice aligns with current SOTA Hebbian architectures \citep{journe2022hebbian}, moving beyond the single-CNN layer hard-WTA approach of \cite{grinberg2019local}. 

For comparison, we also implemented Lagani's architecture \citep{lagani2022comparing}, a padding-free design with fewer channels per layer and 3-5 convolutional layers, enabling systematic evaluation of architectural effects on Hebbian learning and competition dynamics. Detailed specifications are provided in \ref{architectures}.

\subsubsection{Depthwise Separable Convolution}
Our first architectural variant replaced standard convolutions in the Journe architecture with Depthwise Separable Convolutions to better approximate biological visual processing while reducing computational complexity. Furthermore, all our previously detailed Hebbian learning rules and competition mechanisms were incorporated into these convolutions to investigate their potential to produce centre-surround receptive fields.

In the depthwise component, we restricted competition and learning to operate independently within the spatial dimensions of each input channel. This was accomplished by computing activations and weight updates separately for each channel's convolutional filter, preventing cross-channel competition. We then implemented the pointwise component as 1x1 standard convolutions that combined channel-wise features, with Hebbian learning rules and competition operating across the channel dimension. The pointwise layer's weight updates were computed using the same learning rules and competition mechanisms as the depthwise layer, but applied to the channel-wise feature combinations rather than spatial patterns.

\subsubsection{Residual Block}

The second variant incorporated residual connections using Depthwise Separable Convolutions in an inverted bottleneck structure, motivated by the presence of feedback connections in biological neural circuits. Each block consisted of three sequential operations: an initial pointwise convolution that expanded the channel dimension by a fixed factor, followed by a depthwise convolution for spatial feature learning, and a final pointwise convolution that projected features back to the original channel dimension. Skip connections bypassed these three operations, directly connecting the input to the output of each block.

This configuration was designed to match the Journe architecture's feature map dimensions and channel counts at corresponding depths, enabling direct performance comparisons. The channel expansion factor of 4 and block placement were specifically chosen to maintain parameter counts comparable to Journe while preserving the network's representational capacity. The complete architectural specifications, including layer dimensions and block placements, are detailed in \ref{architectures}.



\subsection{Dale's Principle Weights}

To pledge biological realism, all the mdodels were designed to allow an optional adherence to Dale's Principle. We ensured all synaptic weights were initialised and maintained excitatory values. This was achieved by using an absolute value function whenever an operation on weights took place. Weight changes still facilitated both Long-Term Potentiation (LTP) and Long-Term Depression (LTD), as both processes are crucial for plasticity and learning.

\subsection{Analysis and Visualisation Methods}

We implemented various visualisation techniques, including UMAP \citep{mcinnes2018umap}, weight distributions and receptive fields, for qualitative and quantitative analysis of the network dynamics during training. Class separability was assessed through Uniform Manifold Approximation and Projection (UMAP) projections of layer feature embeddings, measuring the clustering quality of different class categories. Weight distributions were analysed using kernel density estimation to track the evolution of synaptic strengths.

To understand learned hierarchical feature representations, we implemented Projected Gradient Ascent (PGA) for receptive field visualisation. This technique optimises an input image I to maximise the activation A of specific neurons according to:
$I_{t+1} = I_t + \eta \frac{\partial A}{\partial I} - \lambda I_t$
where $\eta$ represents the learning rate and $\lambda$ controls L2 regularisation to ensure visual coherence.

\subsection{Configurations}

To systematically evaluate our framework's components and their interactions, we designed and implemented 15 total distinct experimental configurations: 3 of these replicated the results of published SOTA models, 11 which built upon previous Hard-WTA research approaches, and a comparative end-to-end backpropagation trained model. Each configuration represents a specific addition of either architectural design, competition mechanisms, or learning rule, allowing us to isolate and analyse their individual effects on network performance. Out of these configurations, the top four will be selected according to an accuracy criterion and thoroughly assessed.

\subsection{Statistical Analyses}

Post-initialisation behaviour performance of the selected configurations was assessed over 5 different seed runs across the last 10 epochs of the test. Descriptives were computed, including arithmetic mean, median, standard deviation and 99\% confidence intervals, which were calculated using the Welch-Satterthwaite equation to account for potential variance differences. We assessed the normality of the distribution using the Shapiro-Wilk test with a p \textgreater 0.05 normality threshold for both paired models. We carried out pairwise comparisons: for non-normally distributed data using the Friedman (Kruskal-Wallis) test, and for normally distributed data using the ANOVA test. We chose a conservative significance threshold of p \textless0.01.

\section{Experimental Settings}

\subsection{Experimental Configurations: Ablation Study}


\begin{table}[h!]
\centering
\caption{Hebbian Learning Initial Configurations and Features}
\label{tab:hebbian-config}
\small
\renewcommand{\arraystretch}{1.3}
\setlength{\tabcolsep}{4pt}
\begin{tabular*}{\textwidth}{@{\extracolsep{\fill}}>{\raggedright\arraybackslash}p{0.42\textwidth} >{\raggedright\arraybackslash}p{0.18\textwidth} >{\raggedright\arraybackslash}p{0.2\textwidth} >{\centering\arraybackslash}p{0.12\textwidth}@{}}
\toprule
\textbf{Configuration} & \textbf{Architecture} & \textbf{Competition} & \textbf{Hebbian Rule} \\
\midrule
\multicolumn{4}{l}{\textit{Replications of Published SOTA models}} \\
\midrule
SoftWTA & Journe & Soft WTA & GI \\
Lagani-HardWTA & Lagani & Hard WTA & GI \\
Lagani\_Deep-HardWTA & Lagani & Hard WTA & GI \\
\midrule
\multicolumn{4}{l}{\textit{Backpropagation Comparison}} \\
\midrule
Backpropagation & Journe & N/A & BP \\
\midrule
\multicolumn{4}{l}{\textit{Novel Implementations}} \\
\midrule
HardWTA & Journe & Hard WTA & GI \\
HardWTA-BCM & Journe & Hard WTA & BCM \\
Presynaptic/HardWTA & Journe & Hard WTA+Pre & GI \\
Temporal/HardWTA & Journe & Hard WTA+Temp & GI \\
Homeostatic/HardWTA & Journe & Hard WTA+Hom & GI \\
Surr/HardWTA$^*$ & Journe & Hard WTA+Surr & GI \\
Depthwise-HardWTA$^\dagger$ & Journe & Hard WTA & GI \\
Residual-HardWTA$^\ddagger$ & Journe & Hard WTA & GI \\
Dale\_Depth-Surr/HardWTA-BCM$^\S$$^\dagger$ & Journe & Hard WTA+Surr & BCM \\
No-WTA & Journe & None & GI \\
\bottomrule
\end{tabular*}
\vspace{0.5em}
{\small \textit{Note:} WTA: Winner-Take-All, GI: Grossberg Instar, BCM: Bienenstock-Cooper-Munro, BP: Backpropagation, Surr: Lateral Surround Inhibition. Special features: $^*$Surround lateral inhibition, $^\dagger$Depthwise separable convolutions, $^\ddagger$Residual connections, $^\S$Dale's Principle.}
\end{table}

Table \ref{tab:hebbian-config} lists all the experimental setups, outlining the architecture, types of competition mechanisms, variants of the Hebbian learning rule for weight updates, and any extra activations and features utilised. Table \ref{tab:hebbian-purposes} summarises the rationale behind each experimental setup, including the choice of specific competition mechanisms, architectures, or learning rules. 

\begin{table}[h!]
\centering
\caption{Purposes of Initial Experimental Hebbian Learning Configurations}
\label{tab:hebbian-purposes}
\small
\begin{tabular}{@{}p{0.40\textwidth}p{0.55\textwidth}@{}}
\toprule
\textbf{Configuration} & \textbf{Purpose} \\
\midrule
\multicolumn{2}{l}{\textit{Replications of Published SOTA models}} \\
\midrule
SoftWTA & Evaluate SoftHebb research \\
Lagani-HardWTA & Evaluate Lagani Hard-WTA research \\
Lagani\_Deep-HardWTA & Evaluate Lagani Hard-WTA 4-layer research \\
\midrule
\multicolumn{2}{l}{\textit{Backpropagation Comparison}} \\
\midrule
Backpropagation & Evaluate pure backpropagation baseline \\
\midrule
\multicolumn{2}{l}{\textit{Novel Implementations}} \\
\midrule
HardWTA & Evaluate Hard-WTA in network with padding \\
HardWTA-BCM & Compare BCM to Grossberg-Instar rule \\
Presynaptic/HardWTA & Assess impact of presynaptic learning \\
Temporal/HardWTA & Evaluate temporal competition with/without WTA \\
Homeostatic/HardWTA & Evaluate statistical competition with Hard-WTA \\
Surr/HardWTA & Assess effect of surround modulation kernel \\
Depthwise-HardWTA & Evaluate depthwise equivalent of hard-WTA research \\
Residual-HardWTA & Evaluate residual equivalent of hard-WTA research \\
Dale\_Depth-Surr/HardWTA-BCM & Evaluate improved biological realism in a CNN \\
No-WTA & Evaluate basic learning without competition \\
\bottomrule
\end{tabular}
\end{table}

The configurations were built upon two fundamental architectural approaches established in previous research: the Journe CNN architecture from \citep{journe2022hebbian}, and the Lagani CNN architecture from \citep{lagani2022comparing}. The \path{SoftWTA} and \path{Lagani-HardWTA} and/or \path{Lagani_Deep-HardWTA} configurations directly replicate these approaches respectively, serving as benchmarks for our enhanced implementations. 

Building upon these foundations, we developed a series of novel configurations that integrate additional biologically-sound mechanisms or architectural improvements. Unless specified otherwise, all HardWTA experiments were conducted using the Journe architecture, incorporating the cosine similarity activation function and employing the Grossberg Instar update rule. Our configurations explored combinations of Hard-WTA competition with presynaptic competition (\path{Presynaptic/HardWTA}), lateral inhibition (\path{Surr/HardWTA}), homeostatic (\path{Homeostatic/HardWTA}) and temporal (\path{Temporal/HardWTA}) competition. \path{No-WTA} evaluated the system performance with no competition mechanism.

We introduced architectural variants including depthwise separable convolutions (\path{Depthwise-HardWTA}) and residual connections (\path{Residual-HardWTA}) to investigate more efficient and biologically-inspired network structures. Either the Grossberg Instar or BCM rule was employed to adjust the convolutional weights.

The \path{Dale-Depthwise-Surr/HardWTA-BCM} configuration represents our most biologically realistic implementation, uniquely combining three key elements: Dale's Principle constraints on synaptic weights, depthwise separable convolutions, and BCM Hebbian learning with Hard-WTA competition and lateral inhibition.

To establish a direct comparison with traditional deep learning approaches, we included the \path{Journe-Backpropation}
configuration, which implements end-to-end backpropagation training using the same architecture as our Hebbian implementations. This configuration serves as a control, allowing us to evaluate the relative performance of our biologically-inspired learning mechanisms against conventional gradient-based methods.

Lastly, an \path{Optimal-HardWTA} configuration was built which integrates the mechanisms shown to improve performance to serve as the optimal configuration. 

\subsection{Network Training}

The training process for the configurations consisted of two distinct phases for all configurations except the end-to-end backpropagation comparison:

1. \textbf{Unsupervised Feature Learning Phase}: Only convolutional layers were trained for one epoch with a batch size of 64 using Hebbian learning rules. Hard-WTA configurations used a learning rate of 0.1 across all layers, while Soft-WTA used a custom per-layer learning rate schedule as detailed in \citep{journe2022hebbian}. A normal random distribution with a large radius range was used to initialise the SoftWTA Hebbian-CNN weights (detailed in \citep{journe2022hebbian} and \ref{weight init}) as learning only occurred with this specific setup. All Hard-WTA and backpropagation experiments were initialised with the default PyTorch Kaiming Uniform distribution \citep{he2015delving}. ZCA-Whitening was applied only for configurations with Hard-WTA competition mechanisms. Custom visualisation tools were used throughout to analyse the network's representational abilities. After feature learning, convolutional layer weights were frozen.
    
2. \textbf{Supervised Classifier Training Phase}: The single-layer classifier head underwent training using backpropagation over 20 epochs employing the Adam optimizer with a learning rate of 0.001. A scheduler was used to cut the learning rate by half every 2 epochs starting from the 10th epoch, and a dropout rate of 0.5 was applied to avoid overfitting, as performed by \cite{journe2022hebbian}. The classifier layer had as input size the flattened features of the CNN, and has as output size the number of classes (10 for CIFAR-10, MNIST and STL10). Both training and test sets were evaluated at each epoch. The pure backpropagation comparison model was trained end-to-end through the Adam optimiser with a learning rate of 0.001.

For consistent comparisons, all experimental configurations were trained using the same predetermined random seed, while distinct seeds were used across the 5 runs per configurations for the statistical analysis. To enable a quantitative comparison among configurations, we applied standard evaluation metrics on the test set during each epoch of the classifier training phase, namely accuracy, precision, recall, and F1 score.

Accuracy is calculated as:
$$\text{Accuracy} = \frac{TP + TN}{TP + TN + FP + FN}$$

F1 score is calculated as:
$$\text{F1} = \frac{2 \times \text{Precision} \times \text{Recall}}{\text{Precision} + \text{Recall}} = \frac{2TP}{2TP + FP + FN}$$

We report accuracy results in decimal form (e.g., 0.92) in all tables, figures, and statistical analyses, while we use percentage form (e.g., 92\%) in the main text to facilitate comparisons with other papers in the literature.

The CIFAR-10 dataset was trained using both the 3-layer Journe and the 3-4 layer Lagani architectures. Training with different datasets entailed some specific modifications: the use of a greyscale variant of the Journe and Lagani architectures for MNIST, and the incorporation of an extra 4th layer in the Journe architecture for the larger images of the STL-10 dataset and usage of the 4-layer Lagani variant to match the identical architecture of \cite{journe2022hebbian}, as elaborated in \ref{architectures}.

\section{Results} \label{results}

\subsection{Classification Accuracy on CIFAR-10}
\begin{table}[h!]
    \centering
    \caption{Top CIFAR-10 Test Performance Metrics for all Experimental Configurations across 20 Epochs}
    \label{tab:config_performance}
    \small
    \setlength{\tabcolsep}{4pt}
    \begin{tabularx}{\textwidth}{@{}Xcc@{}}
    \toprule
    \textbf{Configuration} & \textbf{Accuracy} & \textbf{F1 Score} \\
    \midrule
    \multicolumn{3}{l}{\textit{Replications of Published SOTA models}} \\
        \midrule
    \rowcolor{gray!15} \textbf{SoftWTA} & \textbf{0.792} & \textbf{0.79} \\
    Lagani-HardWTA & 0.597 & 0.59 \\
    Lagani\_Deep-HardWTA & 0.528 & 0.53 \\
    \midrule
    \multicolumn{3}{l}{\textit{Backpropagation Comparison}} \\
    \midrule
    \textbf{Backpropagation} & \textbf{0.777} & \textbf{0.77} \\
    \midrule
    \multicolumn{3}{l}{\textit{Novel Implementations}} \\
        \midrule
   HardWTA & 0.748 & 0.75 \\
    HardWTA-BCM & 0.753 & 0.75 \\
    Presynaptic/HardWTA & 0.651 & 0.65 \\
    Temporal/HardWTA & 0.746 & 0.74 \\
    Homeostatic/HardWTA & 0.685 & 0.69 \\
    Surr/HardWTA & 0.757 & 0.76 \\
    Depthwise-HardWTA & 0.719 & 0.72 \\
    Residual-HardWTA & 0.747 & 0.75 \\
    Dale\_Depthwise-HardWTA-BCM & 0.673 & 0.67 \\
    No-WTA & 0.199 & 0.600 \\
    \midrule
    \multicolumn{3}{l}{\textit{Soft-WTA with BCM+Lateral Inhibition}} \\
    \midrule
    SoftWTA-Surr-BCM & 0.73 & 0.73 \\
    \midrule
    \multicolumn{3}{l}{\textit{Optimal Hard-WTA Configuration}} \\
    \midrule
    \rowcolor{gray!15} \textbf{Optimal-HardWTA} & \textbf{0.76} & \textbf{0.76} \\
    \bottomrule
    \end{tabularx}
\end{table}

\subsubsection{Replications of Published Research}

Our initial configurations focused on reproducing previous research results. The \path{SoftWTA} configuration, with 5.9M parameters, achieved 79.2\% accuracy and 0.79 F1 score, confirming the benchmark performance reported by \cite{journe2022hebbian}. To establish additional baselines, we implemented the \path{Lagani-HardWTA} configuration, and its deeper 4-CNN layer counterpart \path{Lagani_Deep-HardWTA} (with 0.39M and 0.8M parameters), which achieved 59.7\% and 52.8\% accuracy respectively, close to Amato's research \citep{amato2019hebbian} which reached 59.69\% and 49\% in the 3 and 4-layer architectures.

\subsubsection{Ablation Study}

A significant enhancement in Hard-WTA performance was achieved with the new configurations developed, directly building upon the previously described baselines. With the same number of parameters (5.9M), the \path{HardWTA} configuration achieved 74.8\% accuracy and 0.75 F1 score, marking the first time Hard-WTA performance has surpassed \cite{grinberg2019local} ’s SOTA performance (72.2\%) in a deeper network. This integration of the Journe architecture with Hard-WTA mechanisms yielded a substantial improvement of 15.1\% accuracy over the 3-CNN layer Lagani baseline, and an increase of 9.2\% over the previous 3-CNN layer Hard-WTA SOTA (64.6\%) by \cite{miconi2021hebbian}.

Further biologically inspired enhancements rendered additional Hard-WTA improvements. The \path{Surr/HardWTA} configuration achieved 75.7\% accuracy, further improving our new SOTA performance for Hard-WTA implementations. Alternative learning rules with BCM in \path{HardWTA-BCM} also improved performance to 75.3\%, suggesting further evidence of biological efficiency of BCM.

While scoring slightly lower in accuracy (74.6\%), the temporal competition variant, \path{Temporal/HardWTA}, also contributed to highlight the robustness of our enhanced Hard-WTA framework. Other competition mechanisms showed varying degrees of effectiveness, with the \path{Homeostatic/HardWTA} and \path{Presynaptic/HardWTA} configurations achieving 68.5\% and 65.1\% accuracy respectively. 

Our architectural variants focused on efficiency and biological plausibility. The \path{Depthwise-HardWTA}, which significantly reduced model parameters from 5.9M to 0.9M (a factor of 6.6), and the \path{Residual-HardWTA} with 4.03M parameters, achieved 71.9\% and 74.7\% accuracy respectively. They maintained robust performance while reducing network parameters. The putatively most biologically realistic network, \path{Dale_Depthwise-Surr/HardWTA-BCM}, only achieved 67.3\% accuracy with exclusively centre-surround filters at Layer 2-3 (\ref{centre-surr}). This result suggests that inhibitory weights may be essential for attaining high performance. The significant importance of competition mechanisms was emphasised by the \path{No-WTA} setup, which by lacking a competition mechanism reached only 19.9\% accuracy. However, it is worth mentioning that it learnt features that enabled it to exceed the balanced 10-class baseline random accuracy of 10\%.

\subsubsection{Backpropagation Comparison}

The Journe architecture trained through end-to-end backpropagation (\path{Backpropagation}) achieved 77.7\% accuracy. This comparison demonstrates that our enhanced Hard-WTA Hebbian approaches (75\% accuracy) can successfully approximate traditional gradient-based performance in shallow networks. It should be noted that the advantages of backpropagation become more evident with the addition of more deep layers and prolonged training durations, leading to significant performance improvements.

\subsubsection{Combinatorial Setups: Hard and Soft-WTA Combined with Lateral Inhibition and BCM}

The \path{Optimal-HardWTA} configuration was set up by integrating the most effective, biologically inspired mechanisms that enhanced performance. The configuration built included Hard-WTA, layer-specific kernel parameters for Lateral Inhibition, and the BCM learning rule's \path{theta_decay}, similar to \cite{journe2022hebbian} with per-layer Soft-WTA parameters

For Layer1, the settings were $sigma_e = 1.2$, $sigma_i=1.3$, with a lateral kernel size of 5, \path{theta_decay}$=0.3$, and a learning rate of 0.1. For Layer2, $sigma_e = 1.0$, $sigma_i=1.2$, with a lateral kernel size of 3, \path{theta_decay}$=0.35$, and a learning rate of 0.08. Lastly, for Layer3, the parameters were $sigma_e = 0.8$, $sigma_i=1.1$, a lateral kernel size of 3, \path{theta_decay}$=0.35$, and a learning rate of 0.05.

As hypothesised, the resulting \path{Optimal-HardWTA} configuration achieved the highest values: 76\% accuracy with a F1-Score of 0.76.

Following the success of these combinations, the \path{SoftWTA-Surr-BCM} configuration was evaluated using BCM and Lateral Inhibition alongside soft-WTA and anti-Hebbian updates for non-maximum neurons, to measure if these additional competition mechanisms could be integrated successfully without affecting the intrinsic cross-entropy minimisation of the original formulation by \citep{journe2022hebbian}.

The \path{SoftWTA-Surr-BCM} configuration reached 73\% acccuracy and an F1-score of 0.73, suggesting a decrease of the implicit cross-entropy loss function optimisation of the SoftHebb model \citep{moraitis2022softhebb}.

\subsection{Comparison of Prior Work and New Hard-WTA State-of-the-Art}

A detailed evaluation of our framework’s performance was conducted by comparing the results of our replications of the two published Hebbian SOTA models, \path{SoftWTA} and \path{Lagani-HardWTA}, and the \path{Backpropagation} against our highest performing Hard-WTA model, \path{Optimal-HardWTA}.

The following section presents detailed analyses of these selected configurations, examining their performance through accuracy plots and consistency across multiple random seeds and additional datasets, class clustering capabilities through UMAP embeddings, weight distribution characteristics and learnt receptive field patterns.

\subsubsection{Statistical Analysis of CIFAR-10}

We ran each model configuration (\path{SoftWTA}, \path{Backpropagation}, \path{Optimal-HardWTA}, and \path{Lagani-HardWTA}) 5 times with different random initialisation seeds. This approach allowed us to account for variability introduced by weight initialisation and stochastic training processes, providing a more reliable estimate of each model's performance characteristics.

\begin{table}[ht]
\caption{Performance comparison on CIFAR-10 test dataset (last 10 epochs)}
\label{tab:model_comparison}
\resizebox{\columnwidth}{!}{%
\begin{tabular}{lccccc}
\toprule
\textbf{Model} & \textbf{Mean} & \textbf{Median} & \textbf{99\% CI} & \textbf{Std Dev} & \textbf{Max} \\
\midrule
Backpropagation & 0.752 & 0.754 & [0.747, 0.757] & 0.014 & 0.769 \\
\textbf{Optimal-HardWTA} & \textbf{0.752} & \textbf{0.753} & \textbf{[0.750, 0.755]} & \textbf{0.007} & \textbf{0.758} \\
SoftWTA & 0.775 & 0.781 & [0.768, 0.783] & 0.019 & 0.795 \\
Lagani-HardWTA & 0.597 & 0.600 & [0.593, 0.601] & 0.010 & 0.604 \\
\bottomrule
\end{tabular}%
}
\end{table}

Table~\ref{tab:model_comparison} presents the mean CIFAR-10 performance metrics and statistical comparisons for our learning algorithms across the final 50\% test epochs.

Prior to conducting our statistical comparisons, we performed the Shapiro-Wilk tests to assess normality of our data distributions. The results were as follows: Backpropagation (W=0.978, p=0.466); Optimal-HardWTA (W=0.921, p=0.003); SoftWTA (W=0.875, p=$8.12 \times 10^{-5}$); Lagani-HardWTA (W=0.877, p=$9.10 \times 10^{-5}$). These results led us to reject the hypothesis of normality. Due to the non-normal distribution of the data, we employed the non-parametric Friedman (Kruskal-Wallis) tests for pairwise comparisons.

Our proposed \path{Optimal-HardWTA} model achieved comparable performance metrics to the traditional \path{Backpropagation}. The two model accuracy levels  did not statistically differ ($H(1) = 0.11$, $p = 0.736$).

The existing \path{SoftWTA} approach significantly outperformed our \path{Optimal-HardWTA} model ($H(1) = 35.32$, $p = 2.80 \times 10^{-9}$;), evidencing the effectiveness of this established biologically-plausible alternative. Similarly, the accuracy performace of the \path{SoftWTA} method was significantly superior to \path{Backpropagation} ($H(1) = 33.06$, $p = 8.95 \times 10^{-9}$;).

Our \path{Optimal-HardWTA} model substantially outperformed the \path{Lagani-HardWTA} baseline ($H(1) = 74.26$, $p = 6.84 \times 10^{-18}$; ). \path{Backpropagation} ($H(1) = 74.26$, $p = 6.84 \times 10^{-18}$;) and \path{SoftWTA} ($H(1) = 74.26$, $p = 6.85 \times 10^{-18}$;) also significantly outperformed this baseline model.

We additionally conducted a separate analysis on the final epoch to evaluate the ultimate performance of each model after the learning process was complete. Before performing our statistical analyses on the last epoch, we performed the Shapiro-Wilk test to assess normality of our data distributions using a significance threshold of $\alpha = 0.05$. The results indicated normal distributions for all models: \path{Backpropagation} (W=0.809, p=0.096); \path{Optimal-HardWTA} (W=0.965, p=0.844); \path{SoftWTA} (W=0.950, p=0.738); \path{Lagani-HardWTA} (W=0.905, p=0.441). Shapiro-Wilk test analyses did not allow us to reject the default hypothesis of normal distribution of data. Hence, normality was assumed, and a series of ANOVA tests were conducted to compare pairs of configurations.

Our proposed \path{Optimal-HardWTA} model (75.6\% mean accuracy performance value during the last epoch ) achieved comparable performance metrics to the traditional \path{Backpropagation} approach (76.9\%) ($F(1,8) = 6.43$, $p = 0.035$), not differing significantly from this traditional gradient-based method.

The \path{SoftWTA} model (79.45\%) significantly outperformed the \path{Backpropagation} method ($F(1,8) = 23.61$, $p = 0.001$;). Likewise, the \path{SoftWTA} model significantly performed better than our \path{Optimal-HardWTA} approach ($F(1,8) = 451.25$, $p = 2.54 \times 10^{-8}$;), highlighting the superior performance of this established method over our proposed biologically-plausible alternative.

Both our \path{Optimal-HardWTA} model and the \path{Backpropagation} method substantially outperform the \path{Lagani-HardWTA} baseline (60.3\%). These differences were significantly reliable ($F(1,8) = 4090.46$, $p = 3.97 \times 10^{-12}$; and $F(1,8) = 936.93$, $p = 1.41 \times 10^{-9}$, respectively). The \path{SoftWTA} model also significantly outperformed \path{Lagani-HardWTA} ($F(1,8) = 7260.67$, $p = 4.01 \times 10^{-13}$).

\begin{figure}[h!]
\centering
\includegraphics[width=\textwidth]{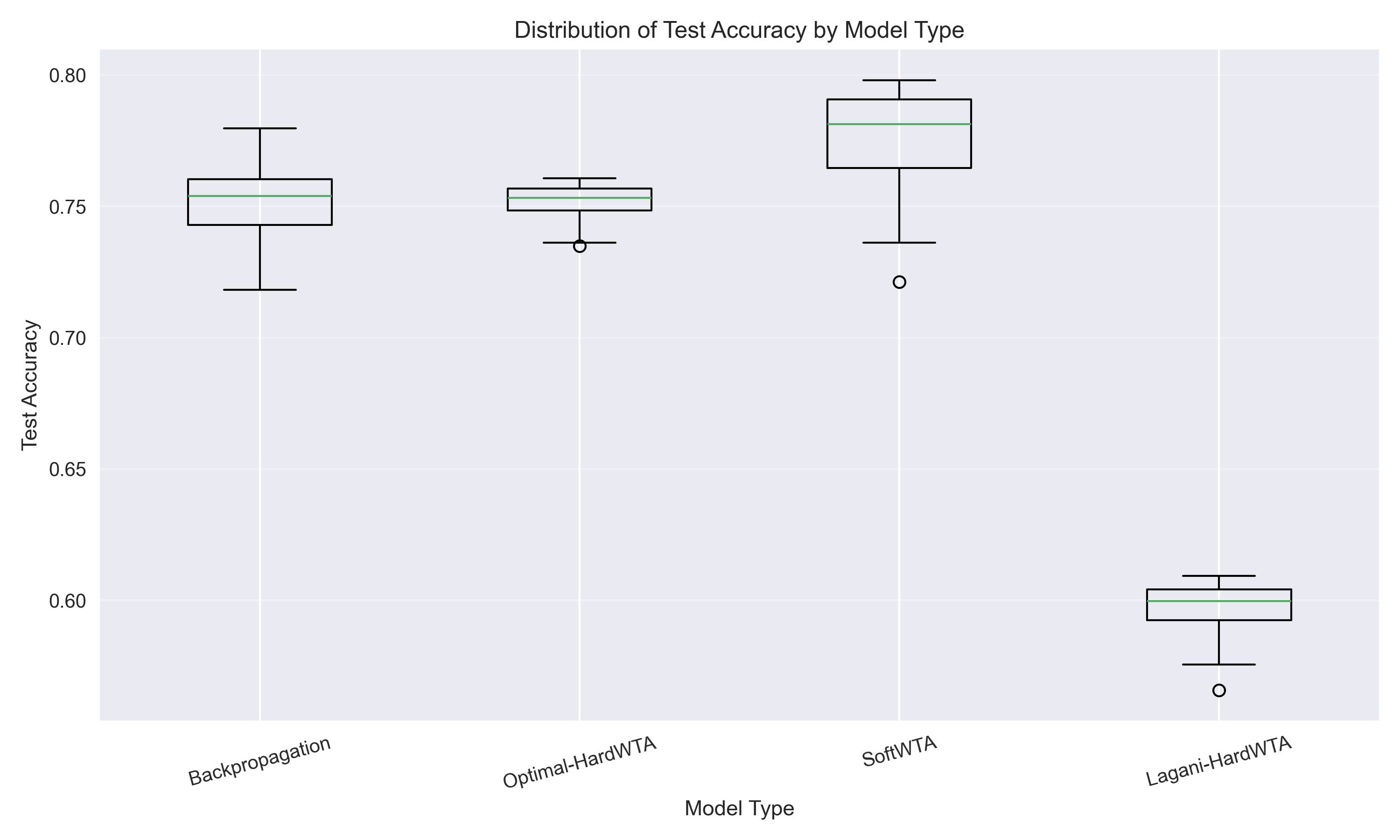}
\caption{Boxplot of CIFAR-10 Test Accuracy by Configuration across different random seeds}
\label{fig:boxplot_accuracy}
\end{figure}

Figure~\ref{fig:boxplot_accuracy} presents the boxplot distribution of mean test accuracy across multiple runs with different random seeds for each model. The \path{SoftWTA} model displayed the highest median accuracy (77.5\%) with a wider interquartile range, indicating strong overall performance but with some variability across different initialisations. One notable outlier at the lower end suggests occasional suboptimal convergence.

Our \path{Optimal-HardWTA} model exhibits consistency, with the narrowest interquartile range among all models and lowest standard deviation (0.007). Its median performance (75.3\%) closely aligns with the \path{Backpropagation} approach (75.4\%), visually reinforcing our conclusion that these models perform comparably. However, the backpropagation variant shows greater variability, with a wider spread between the first and third quartiles.

The \path{Lagani-HardWTA} model is clearly separated from the other approaches, with substantially lower accuracy (median 60\%) and no overlap with the performance distributions of the other models. This visual separation is consistent with the reported significance differences.

\subsubsection{CIFAR-10 Accuracy Dynamics}

\begin{figure}[h!]
\centering
\resizebox{0.65\textwidth}{!}{%
\begin{minipage}[b]{0.65\textwidth}
\includegraphics[width=\textwidth]{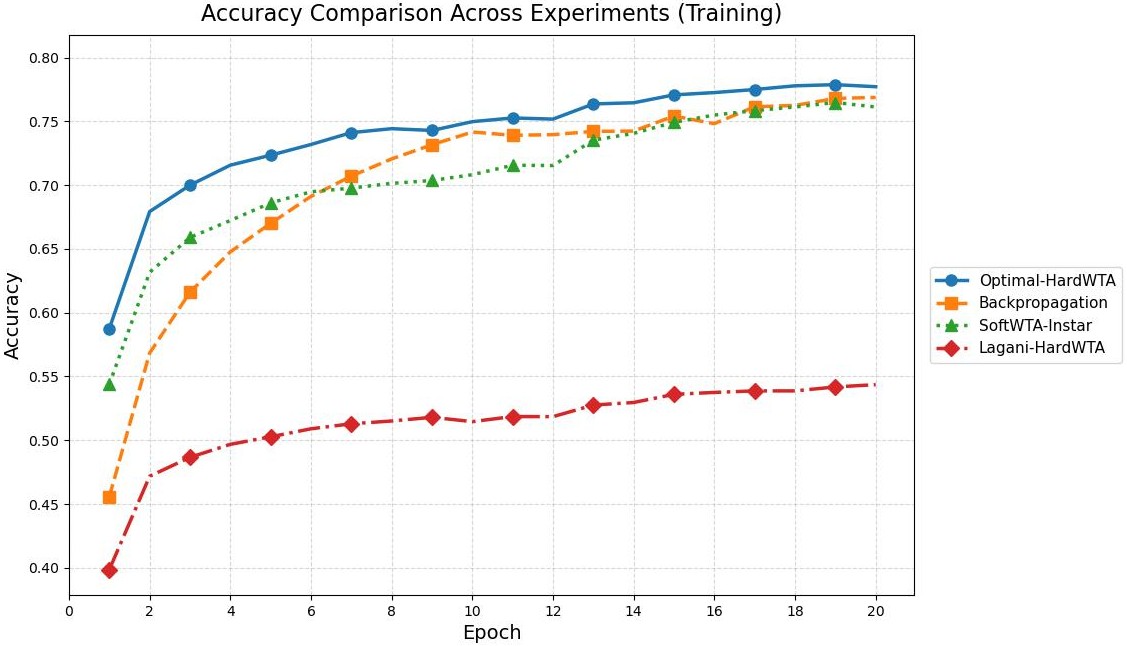}
  \centering
\end{minipage}
}
\caption{Accuracy values during CIFAR-10 training for Lagani-HardWTA (red-rhombus-dot-dash line), SoftWTA (green-triangle-dotted line), Backpropagation (orange-square-broken line), and Optimal-HardWTA (blue-circle-solid line)}
\label{fig:acc-train}
\end{figure}

\begin{figure}[h!]
\centering
\resizebox{0.65\textwidth}{!}{%
\begin{minipage}[b]{0.65\textwidth}
\includegraphics[width=\textwidth]{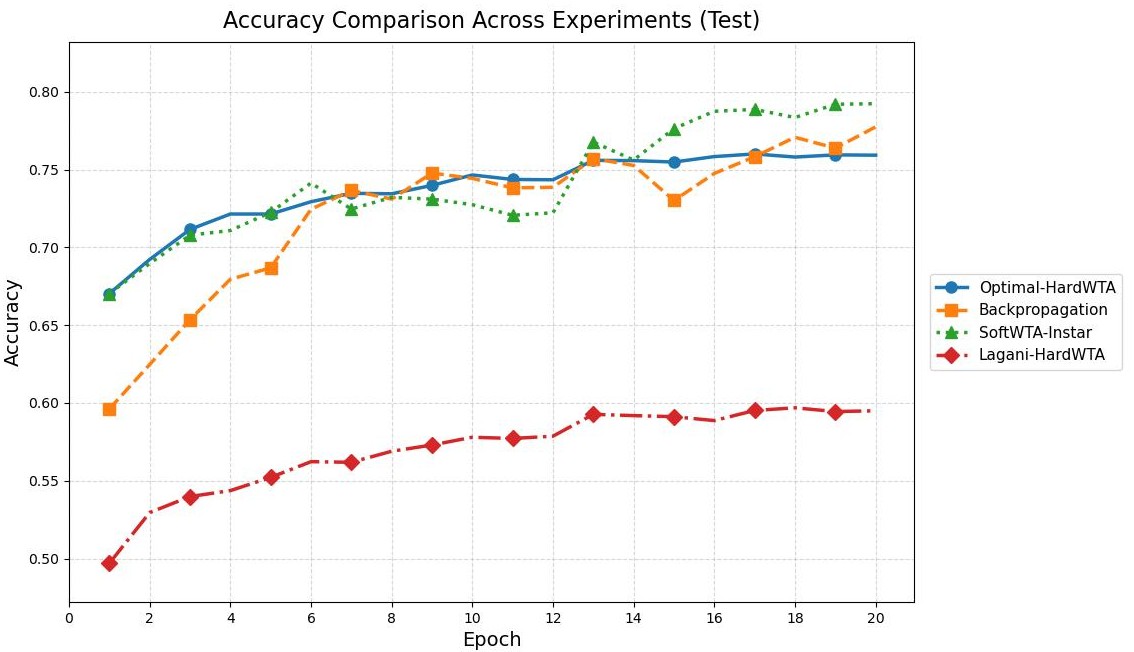}
  \centering
\end{minipage}
}
\caption{Accuracy values during CIFAR-10 test forLagani-HardWTA (red-rhombus-dot-dash line), SoftWTA (green-triangle-dotted line), Backpropagation (orange-square-broken line), and Optimal-HardWTA (blue-circle-solid line)}
\label{fig:acc-test}
\end{figure}

Figure \ref{fig:acc-train} displays accuracy per epoch during training and Figure \ref{fig:acc-test} the test accuracy for the \path{Optimal-HardWTA} (solid blue line), \path{Backpropagation} (dashed orange line), \path{SoftWTA} (dotted green line), and \path{Lagani-HardWTA} (red dash-dotted line).

Training and Test accuracy were remarkably higher and converged faster in our \path{Optimal-HardWTA} than the previous Hard-WTA research \path{Lagani-HardWTA}. Moreover, the accuracy achieved by both \path{Optimal-HardWTA} and \path{SoftWTA} variants were comparable throughout training (Figure \ref{fig:acc-train}), with both Hebbian WTA approaches surpassing \path{Backpropagation} at each epoch in both higher accuracy and convergence, with a significant higher initial accuracy for \path{Optimal-HardWTA}. A similar pattern was observed during test (Figure \ref{fig:acc-test}), although the final accuracy level of the \path{SoftWTA} was slightly higher than other configurations.

\subsubsection{Extension study to other datasets}
\label{sec:ablation}

A study was conducted on both MNIST and STL-10 datasets to assess generalisation across different data types and scales. All reported results are for the last 10 epochs during the testing phase of 5 random seed runs.

Table~\ref{tab:ablation_mnist} shows the comparative performance of all model variants on the MNIST dataset.

\begin{table}[htbp]
\caption{Performance comparison on MNIST test dataset (last 10 epochs)}
\label{tab:ablation_mnist}
\resizebox{\columnwidth}{!}{%
\begin{tabular}{lccccc}
\hline
\textbf{Model} & \textbf{Mean} & \textbf{Median} & \textbf{99\% CI} & \textbf{Std Dev} & \textbf{Max} \\
\hline
Backpropagation & 0.9853 & 0.9859 & [0.9843, 0.9863] & 0.0027 & 0.9884 \\
\textbf{Optimal-HardWTA} & \textbf{0.9829} & \textbf{0.9833} & \textbf{[0.9820, 0.9837]} & \textbf{0.0022} & \textbf{0.9850} \\
SoftWTA & 0.9787 & 0.9806 & [0.9767, 0.9808] & 0.0053 & 0.9830 \\
Lagani-HardWTA & 0.9468 & 0.9473 & [0.9457, 0.9479] & 0.0029 & 0.9493 \\
\hline
\end{tabular}%
}
\end{table}

We performed Shapiro-Wilk tests to assess normality of our data distributions. The results forced us to reject the normal distribution hypothesis: Backpropagation (W=0.885, $p=1.62 \times 10^{-4}$); Optimal-HardWTA (W=0.861, $p=3.13 \times 10^{-5}$); SoftWTA (W=0.755, $p=9.22 \times 10^{-8}$); Lagani-HardWTA (W=0.955, p=0.057). Thus, we carried out non-parametric Friedman (Kruskal-Wallis) tests for pairwise comparisons.

Our proposed \path{Optimal-HardWTA} model achieved strong performance metrics (mean 98.29\%). However, the small difference with the traditional \path{Backpropagation} approach (mean 98.53\%) was statistically reliable, as confirmed by the Friedman test ($H(1) = 27.25$, $p = 1.79 \times 10^{-7}$). While traditional gradient-based methods maintain a slight advantage on the MNIST dataset, our biologically-inspired learning rule approaches comparable performance while maintaining biological plausibility. Notably, the \path{Optimal-HardWTA} approach demonstrates lower variability (std dev: 0.0022 vs. 0.0027), suggesting more stable learning dynamics across training runs.

It is worth noticing that unlike with the CIFAR-10 dataset, our \path{Optimal-HardWTA} model outperformed the \path{SoftWTA} approach (mean 97.87\%) ($H(1) = 33.15$, $p = 8.53 \times 10^{-9}$). This result highlights the effectiveness of our approach over previous biologically-plausible alternatives. Both our model and the \path{Backpropagation} method significantly outperformed the \path{SoftWTA} model, with backpropagation showing the largest performance advantage ($H(1) = 54.63$, $p = 1.45 \times 10^{-13}$).

\path{Optimal-HardWTA} model outperformed the \path{Lagani-HardWTA} baseline (mean 94.68\%). These differences were significantly reliable ($H(1) = 74.28$, $p = 6.78 \times 10^{-18}$). Similarly, both \path{Backpropagation} ($H(1) = 74.27$, $p = 6.80 \times 10^{-18}$) and \path{SoftWTA} ($H(1) = 74.27$, $p = 6.80 \times 10^{-18}$) also significantly outperform this baseline model.

We additionally conducted a separate analysis on the final epoch to evaluate the ultimate performance of each model after the learning process was complete. The results did not allow us to reject the null normal distribution hypothesis: Backpropagation (W=0.833, p=0.148); Optimal-HardWTA (W=0.894, p=0.377); SoftWTA (W=0.912, p=0.482); Lagani-HardWTA (W=0.981, p=0.942). Hence, normality was assumed, and a series of ANOVA tests were conducted to compare pairs of configurations.

Our proposed \path{Optimal-HardWTA} model (98.47\% mean accuracy performance value during the last epoch) achieved comparable performance metrics to the traditional \path{Backpropagation} approach (98.59\%), and no statistic differences were found between these two configurations ($F(1,8) = 0.93$, $p = 0.363$). These results suggest that the \path{Optimal-HardWTA} model achieved a performance that could not be reliable distinguished from the traditional gradient-based methods at the conclusion of training on the MNIST dataset while maintaining biological plausibility.

The \path{Backpropagation} method performed similarly to the \path{SoftWTA} model (98.26\%) in the final epoch ($F(1,8) = 6.24$, $p = 0.037$). Our \path{Optimal-HardWTA} model significantly outperformed the \path{SoftWTA} approach at the final epoch ($F(1,8) = 17.64$, $p = 0.003$). These results support a superior performance of our method over previous biologically-plausible alternatives.

Both our \path{Optimal-HardWTA} model and the \path{Backpropagation} method substantially outperform the \path{Lagani-HardWTA} baseline (94.8\%) in the final epoch. The difference in performance was statistically reliable ($F(1,8) = 3038.66$, $p = 1.30 \times 10^{-11}$; and $F(1,8) = 711.21$, $p = 4.20 \times 10^{-9}$, respectively). The \path{SoftWTA} model also significantly outperformed this Hard-WTA baseline ($F(1,8) = 1819.02$, $p = 1.01 \times 10^{-10}$).

\begin{table}[h!]
\caption{Performance comparison on STL-10 test dataset (last 10 epochs)}
\label{tab:ablation_stl10}
\resizebox{\columnwidth}{!}{%
\begin{tabular}{lccccc}
\hline
\textbf{Model} & \textbf{Mean} & \textbf{Median} & \textbf{99\% CI} & \textbf{Std Dev} & \textbf{Max} \\
\hline
Backpropagation & 0.5821 & 0.5871 & [0.5712, 0.5930] & 0.0289 & 0.6185 \\
\textbf{Optimal-HardWTA} & \textbf{0.6949} & \textbf{0.7023} & \textbf{[0.6880, 0.7018]} & \textbf{0.0182} & \textbf{0.7091} \\
SoftWTA & 0.6603 & 0.6823 & [0.6435, 0.6771] & 0.0443 & 0.6931 \\
Lagani-HardWTA & 0.5296 & 0.5364 & [0.5203, 0.5388] & 0.0243 & 0.5521 \\
\hline
\end{tabular}%
}
\end{table}

To test the generalisation capabilities on more complex data, we also evaluated all models on the STL-10 dataset. Table~\ref{tab:ablation_stl10} presents these results.

We performed Shapiro-Wilk tests to assess normality of our data distributions. The results compelled us to reject the hypothesis of normality: Backpropagation (W=0.958, $p = 0.070$); Optimal-HardWTA (W=0.797, $p = 7.62 \times 10^{-7}$); SoftWTA (W=0.761, $p = 1.24 \times 10^{-7}$); Lagani-HardWTA (W=0.932, $p = 0.007$). Hence, we conducted the non-parametric Friedman (Kruskal-Wallis) tests for pairwise comparisons.
 
The performance of our \path{Optimal-HardWTA} model was the highest for this dataset (mean 69.49\%), significantly outperforming all other approaches including the backpropagation method (mean 58.21\%). These superior performance over \path{Backpropagation}  was supported as confirmed by the Friedman test ($H(1) = 74.02$, $p = 7.72 \times 10^{-18}$). In addition, our \path{Optimal-HardWTA} approach showed the lowest variability (std dev: 0.0182), indicating more stable learning dynamics across training runs.

\path{Optimal-HardWTA} model also significantly outperformed the \path{SoftWTA} variant on STL-10 ($H(1) = 34.66$, $p = 3.92 \times 10^{-9}$), further suggesting better model accuracy or our approach on complex datasets. The \path{SoftWTA} model showed intermediate performance (mean: 66\%), significantly outperforming both the backpropagation implementation ($H(1) = 53.00$, $p = 3.34 \times 10^{-13}$) and the \path{Lagani-HardWTA} method (mean 52.96\%) ($H(1) = 67.19$, $p = 2.47 \times 10^{-16}$). Despite exhibiting the highest variability (std dev: 0.0443) among all tested models, \path{SoftWTA} still maintains a clear statistical advantage over backpropagation on this more complex dataset. At this point it should be noted that the reported 76\% accuracy on STL-10 by \cite{journe2022hebbian} was achieved through longer training epochs (up to 100).

Our proposed \path{Optimal-HardWTA} model substantially outperformed the \path{Lagani-HardWTA} baseline on the STL-10 dataset, with dramatic improvement observed ($H(1) = 74.26$, $p = 6.84 \times 10^{-18}$). The traditional \path{Backpropagation} approach also significantly outperformed the \path{Lagani-HardWTA} baseline ($H(1) = 51.45$, $p = 7.33 \times 10^{-13}$).

To assess the final performance on STL-10, we conducted a further comparative analysis with the last epoch accuracy values. Shapiro-Wilk tests gave no support to reject the normality of the data on the last epoch: Backpropagation (W=0.953, $p = 0.755$); Optimal-HardWTA (W=0.987, $p = 0.969$); SoftWTA (W=0.951, $p = 0.746$); Lagani-HardWTA (W=0.912, $p = 0.483$). Therefore, ANOVA tests were conducted to compare performance.

Our proposed \path{Optimal-HardWTA} model (70.8\% mean accuracy performance value during the last epoch) significantly outperformed the traditional \path{Backpropagation} approach (57.63\%) ($F(1,8) = 360.33$, $p = 6.14 \times 10^{-8}$). The \path{SoftWTA} model (69\%) also significantly outperformed the \path{Backpropagation} method in the final epoch ($F(1,8) = 229.50$, $p = 3.57 \times 10^{-7}$), further supporting the effectiveness of biologically-plausible learning approaches on challenging visual tasks.

Our \path{Optimal-HardWTA} model maintained its significant advantage over the \path{SoftWTA} variant in the final epoch ($F(1,8) = 27.22$, $p = 8.05 \times 10^{-4}$). Interestingly, the traditional \path{Backpropagation} approach did not significantly outperform the \path{Lagani-HardWTA} baseline (57.6\%) in the final epoch ($F(1,8) = 10.02$, $p = 0.013$).

Our proposed \path{Optimal-HardWTA} model substantially outperformed the \path{Lagani-HardWTA} baseline in the final epoch, with dramatic improvement observed ($F(1,8) = 1098.65$, $p = 7.49 \times 10^{-10}$). Similarly, the accuracy performance of \path{SoftWTA} model was also significantly better than the \path{Lagani-HardWTA} baseline in the final epoch ($F(1,8) = 625.50$, $p = 6.99 \times 10^{-9}$). The results during this final epoch on STL-10, strongly support the superior performance of our biologically-plausible approach, which not only outperformed existing biologically-plausible alternatives but also exceeds the capabilities of traditional gradient-based methods on complex visual recognition tasks.

\subsubsection{UMAP Class Clustering Embeddings}

The UMAP, weight distributions, and PGA receptive fields were all executed using the identical fixed seed during the CIFAR-10 trial.

\begin{figure}[h!]
  \centering
  \resizebox{0.999\textwidth}{!}{%
    \begin{minipage}[b]{0.499\textwidth}
      \includegraphics[width=\textwidth]{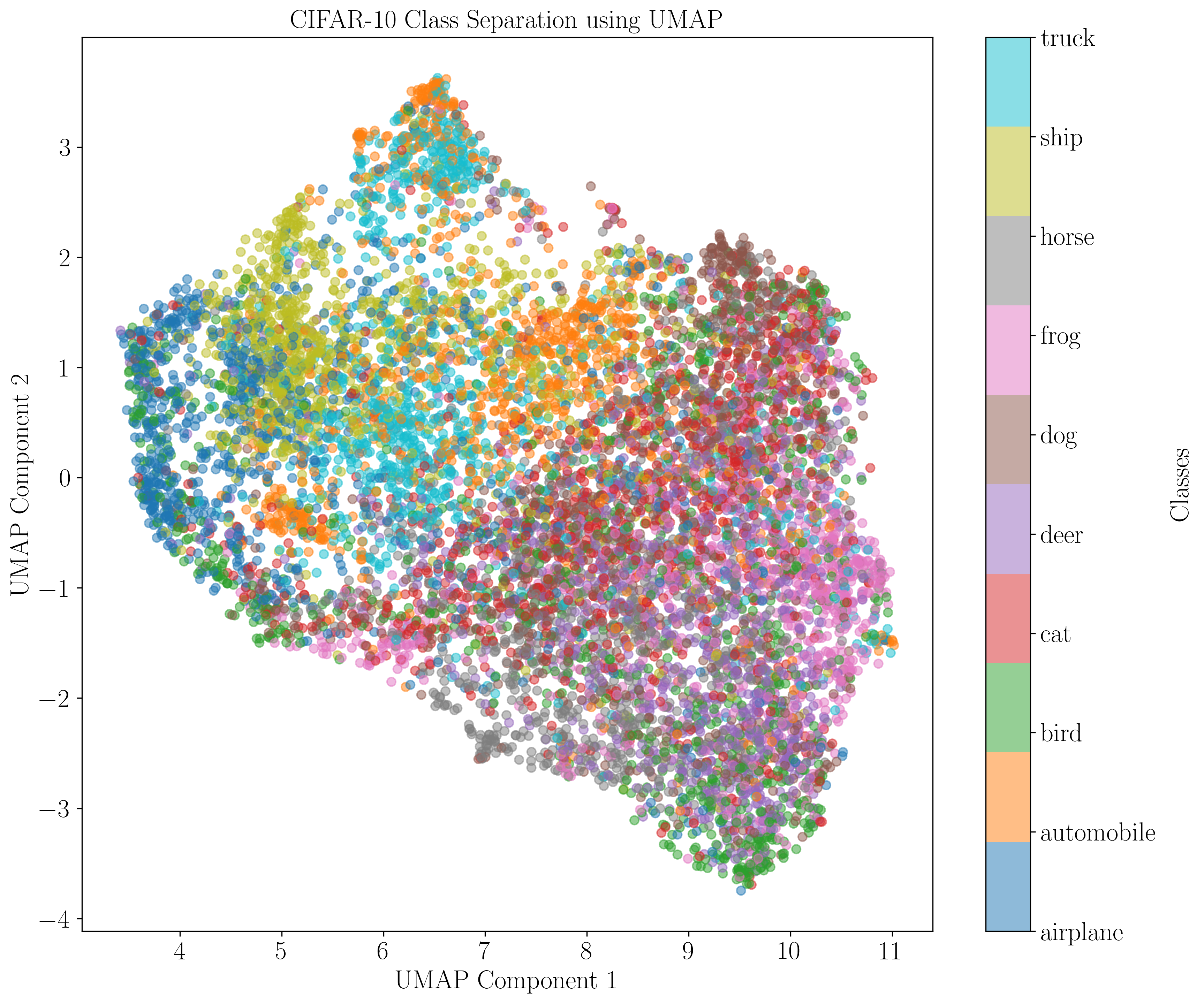}
      \centering
      \text{(A) SoftWTA}
    \end{minipage}
    \hfill
    \begin{minipage}[b]{0.499\textwidth}
      \includegraphics[width=\textwidth]{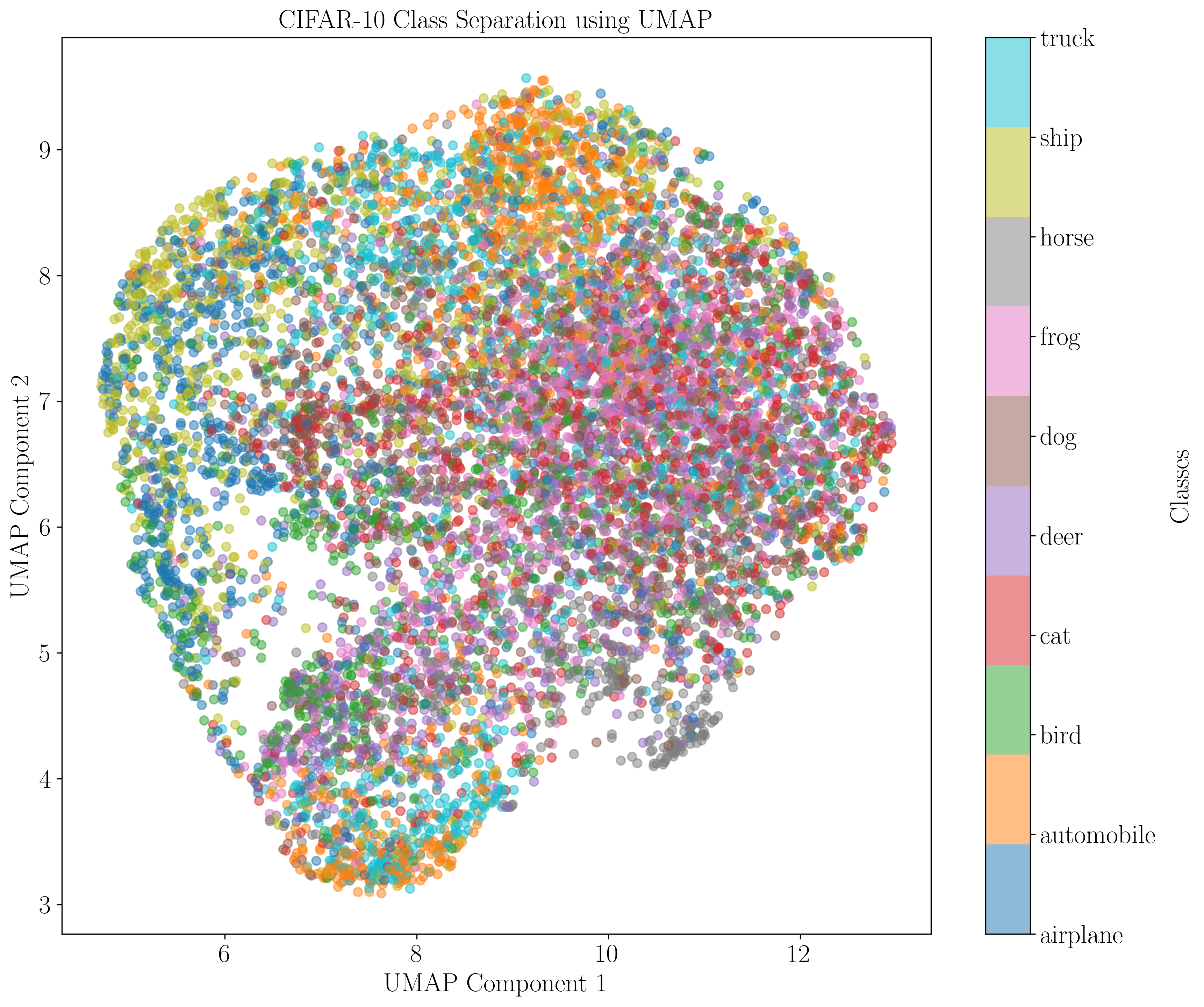}
      \centering
      \text{(B) Lagani-HardWTA}
    \end{minipage}
  }
  \caption{UMAP projection of CIFAR-10 class clusters from unsupervised feature extraction of the \textbf{SoftWTA} and \textbf{Lagani-HardWTA} configurations.}
  \label{fig:umap}
\end{figure}

\begin{figure}[h!]
  \centering
  \resizebox{0.999\textwidth}{!}{%
    \begin{minipage}[b]{0.499\textwidth}
      \includegraphics[width=\textwidth]{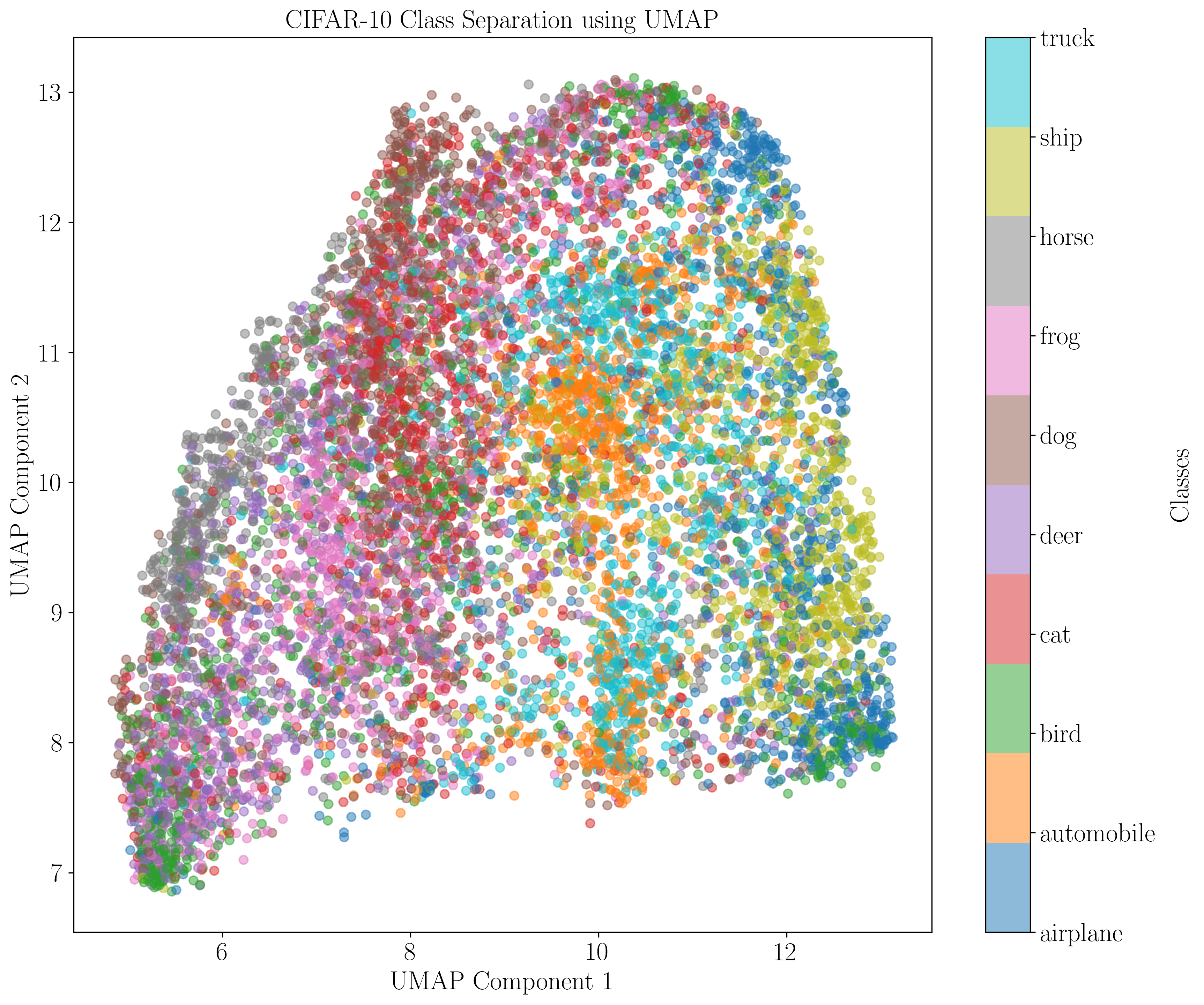}
      \centering
      \text{(A) Optimal-HardWTA}
    \end{minipage}
    \hfill
    \begin{minipage}[b]{0.499\textwidth}
      \includegraphics[width=\textwidth]{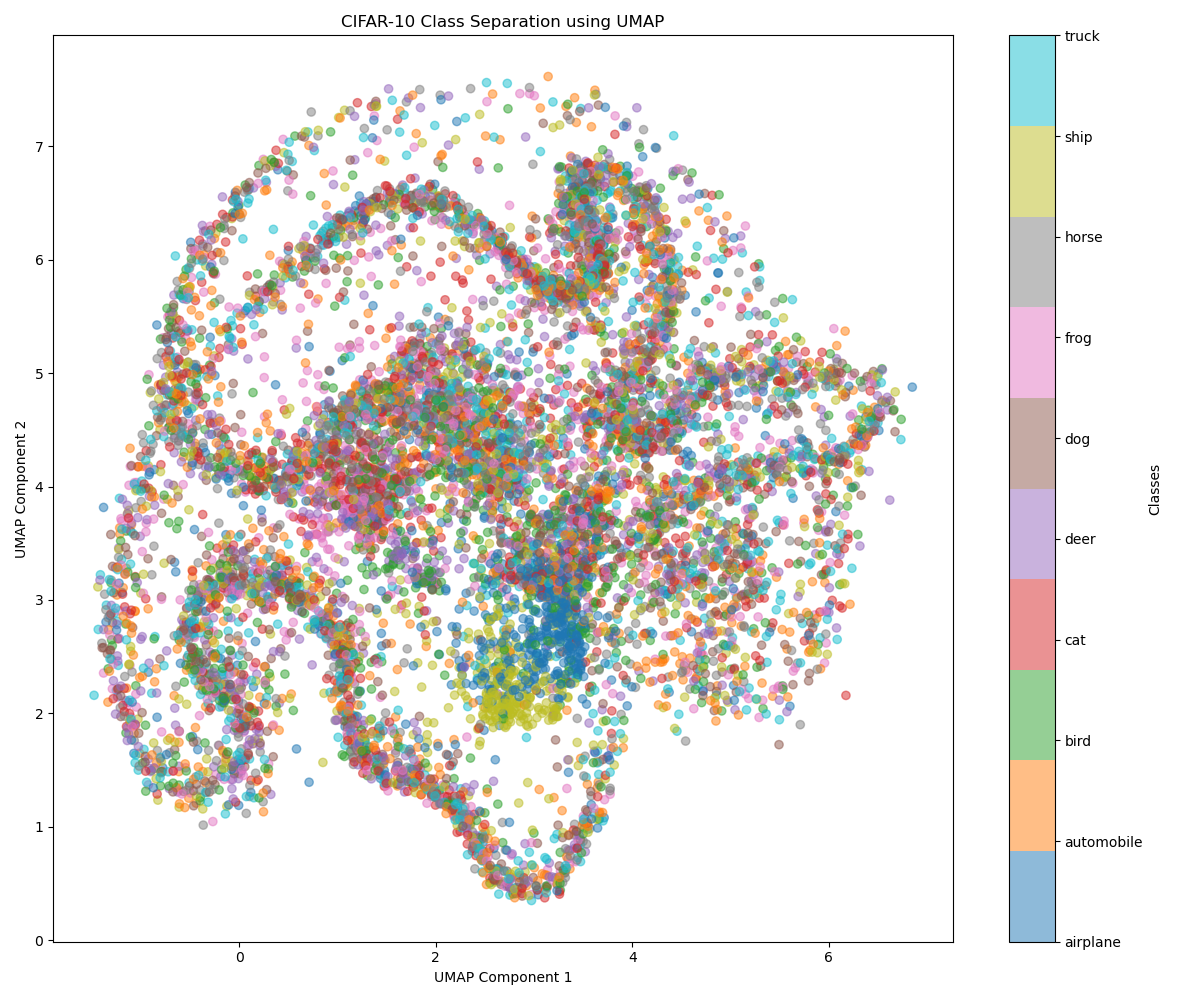}
      \centering
      \text{(B) Backpropagation}
    \end{minipage}
  }
  \caption{UMAP projection of CIFAR-10 class clusters from unsupervised feature extraction of the \textbf{Optimal-HardWTA} and \textbf{Backpropagation} configurations.}
  \label{fig:umap2}
\end{figure}

The UMAP embeddings at the final layer (Figures \ref{fig:umap} and \ref{fig:umap2}) revealed distinct patterns in feature organization. In these visualisations, dots of the same colour represent class labels, with Hebbian configurations showing clustering tendencies, where instances of the same class gather together and are separated from other classes.

The current Hebbian SOTA \path{SoftWTA} (Figure \ref{fig:umap} A) demonstrated highly effective class separation, with clearly delineated clusters showing minimal overlap between classes. Some cross-clustering occurred at cluster boundaries, but the overall organisation suggested effective hierarchical learning with semantically related classes (e.g., automobile in orange and truck in cyan) positioned in proximity to each other while semantically different classes (e.g., ship in yellow and cat in red) located at distant positions.

The previous Hard-WTA benchmark \path{Lagani-HardWTA} (Figure \ref{fig:umap} B) showed weaker clustering characteristics, with less defined boundaries and significant overlap between class clusters. This more diffuse distribution pattern indicates limited ability to form discriminative feature representations, explaining its lower classification performance.

Our enhanced Hard-WTA framework, \path{Optimal-HardWTA} (Figure \ref{fig:umap2} A), achieved clustering quality approaching that of \path{SoftWTA}. The visualisation reveals well-defined semantic clusters while maintaining natural class boundaries, validating that our enhanced competition mechanisms enable more effective feature learning than previous Hard-WTA approaches.

Intriguingly, while \path{Backpropagation} (Figure \ref{fig:umap2} B) achieved high classification accuracy, its UMAP embedding shows a radically different organisation of the learnt feature space. Unlike the Hebbian models which form clear cluster boundaries, the backpropagation model creates a more continuous manifold structure where classes flow into each other with no distinct separation. 

This striking difference in representation structure, despite similar performance metrics, suggests that backpropagation and Hebbian learning discover fundamentally different solutions to the classification task. The backpropagation model appeared to learn non-linear discriminative features for optimisation, which create high-dimensional decision boundaries that do not necessarily cluster visually, rather than the discrete clustering preferred by competitive Hebbian learning. The findings suggest that Hebbian learning offers greater interpretability and explainability compared to backpropagation.

\subsubsection{Weight Distributions}

\begin{figure}[h!]
\centering
\resizebox{0.999\textwidth}{!}{%
\begin{minipage}[b]{0.49\textwidth}\includegraphics[width=\textwidth]{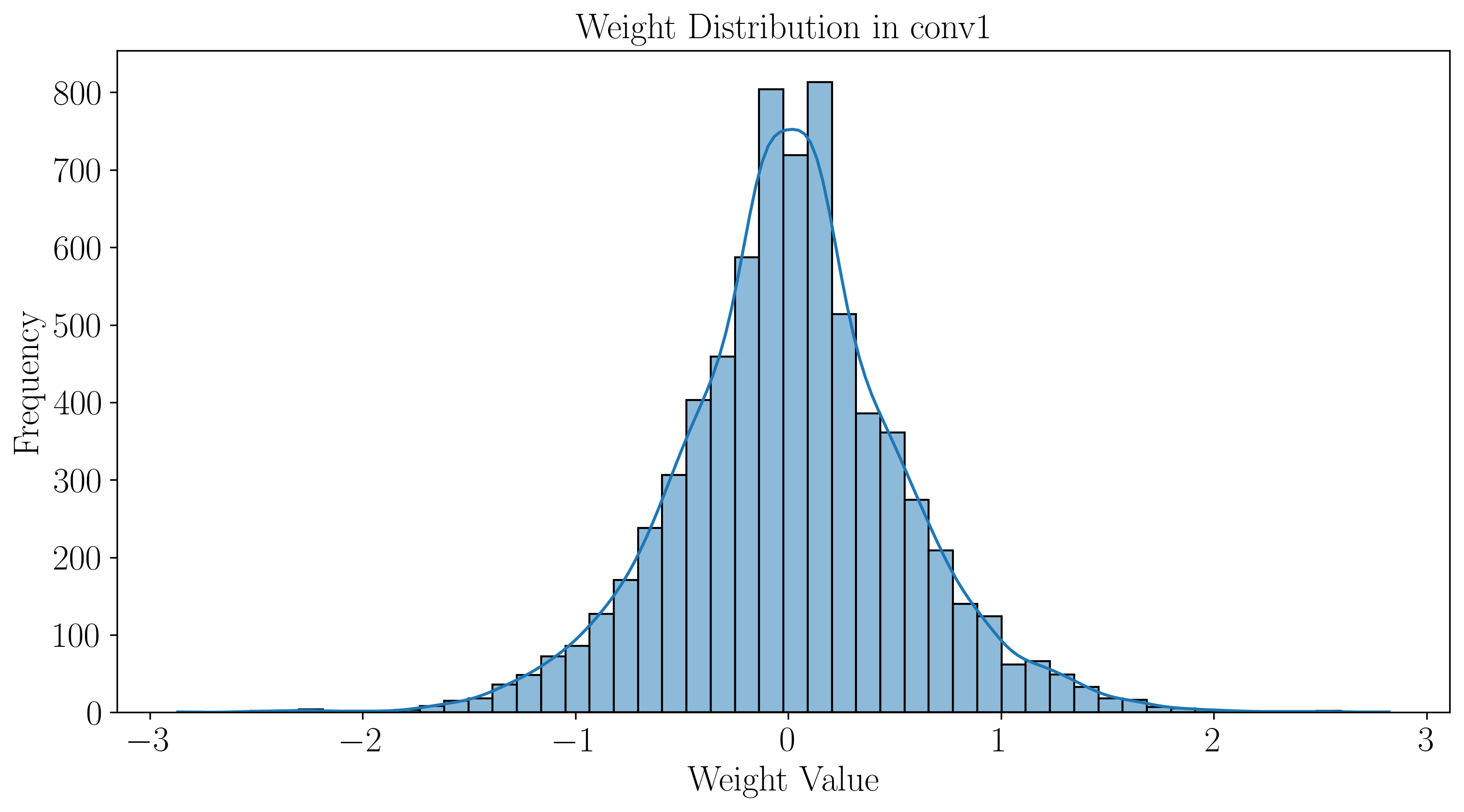}
   \centering
  \textbf{(A) SoftWTA}
\end{minipage}
\hfill
\begin{minipage}[b]{0.49\textwidth}
\includegraphics[width=\textwidth]{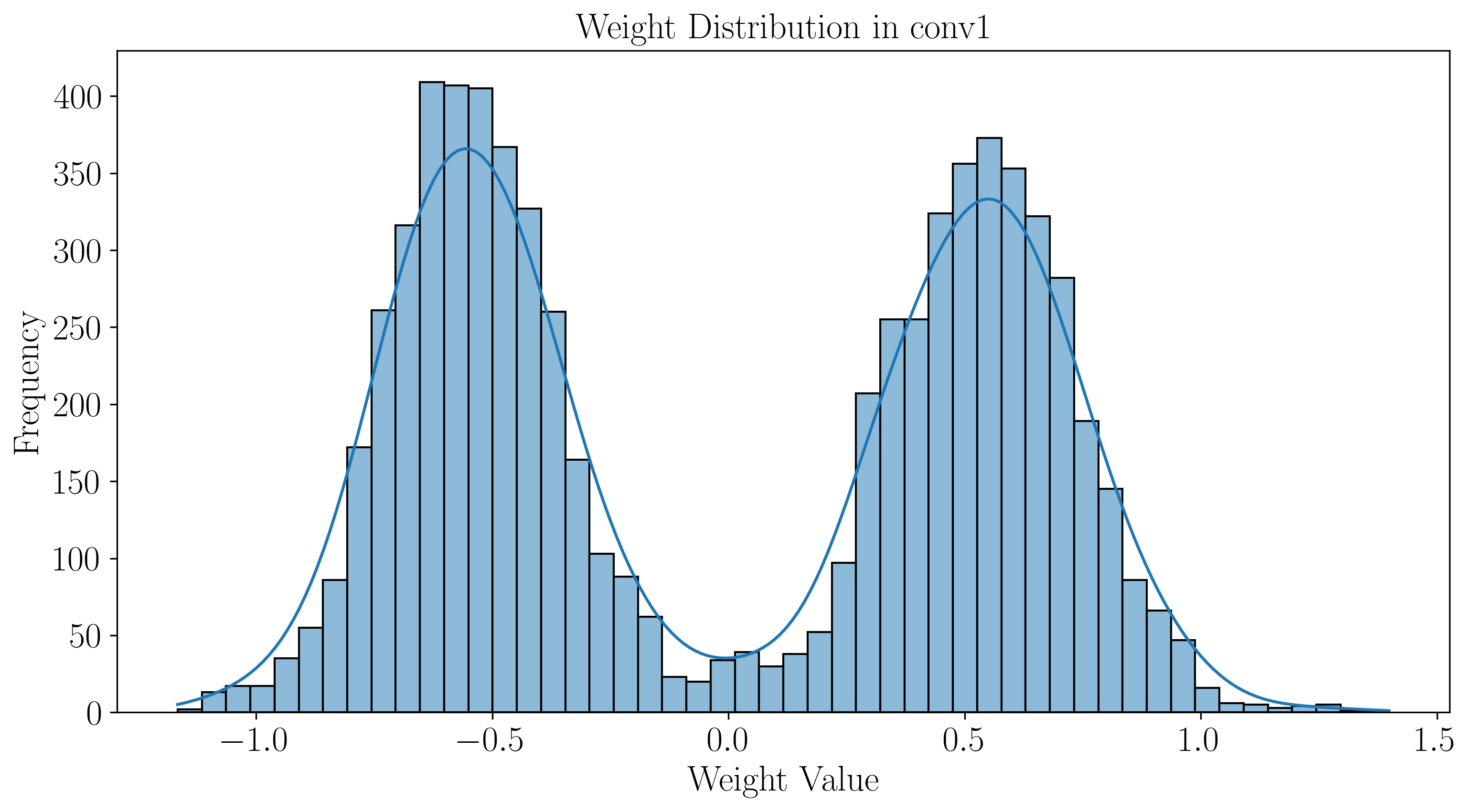}
  \centering
  \textbf{(B) Lagani-HardWTA}
\end{minipage}
}
\caption{Weight distribution of \textbf{SoftWTA} and \textbf{Lagani\_short-Hard/Cos-Instar} configurations for CIFAR-10 at Layer 1.}
\label{fig:weights-dist1}
\end{figure}

\begin{figure}[h!]
\centering
\resizebox{0.999\textwidth}{!}{%
\begin{minipage}[b]{0.49\textwidth}\includegraphics[width=\textwidth]{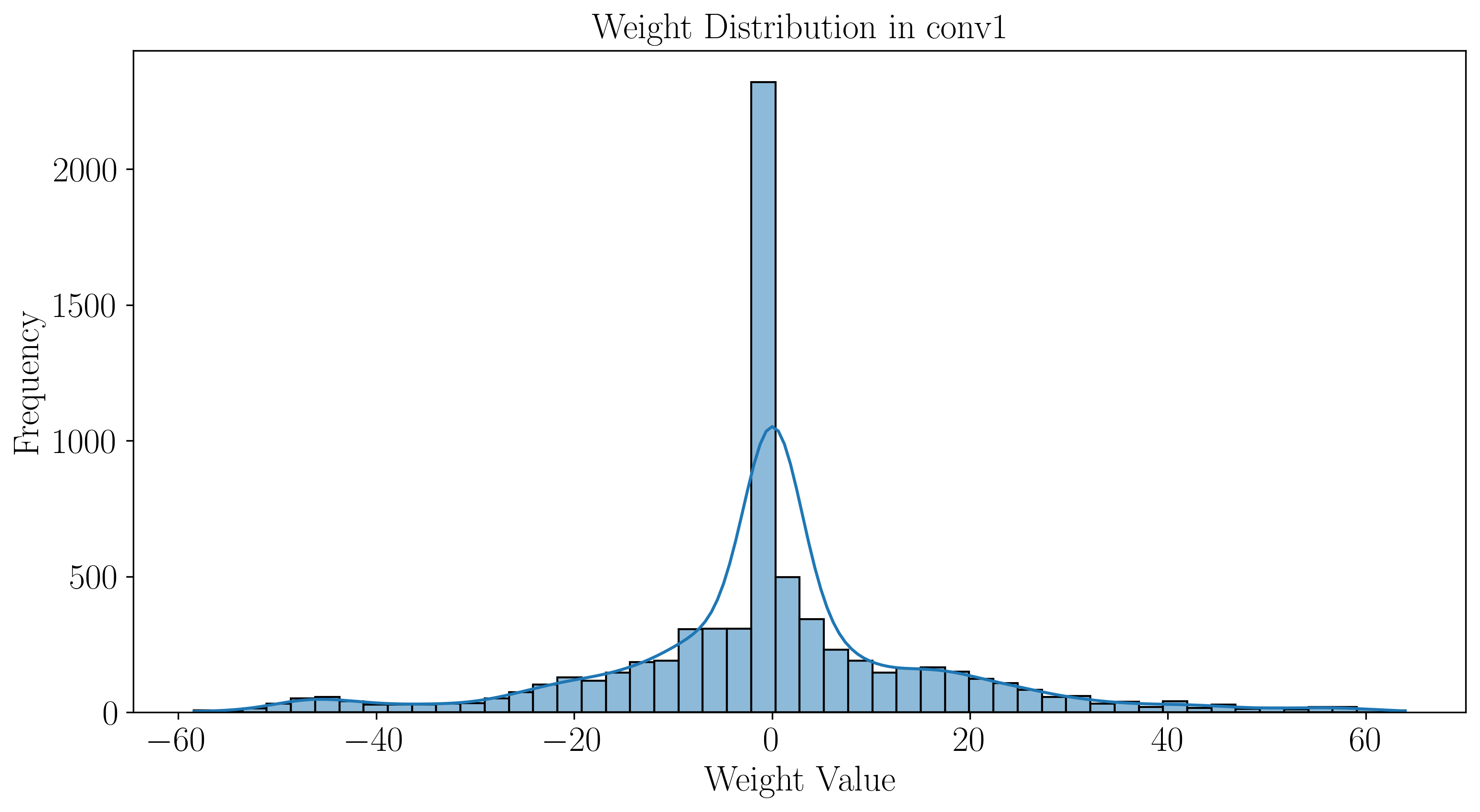}
   \centering
  \textbf{(A) Optimal-HardWTA}
\end{minipage}
\hfill
\begin{minipage}[b]{0.49\textwidth}
\includegraphics[width=\textwidth]{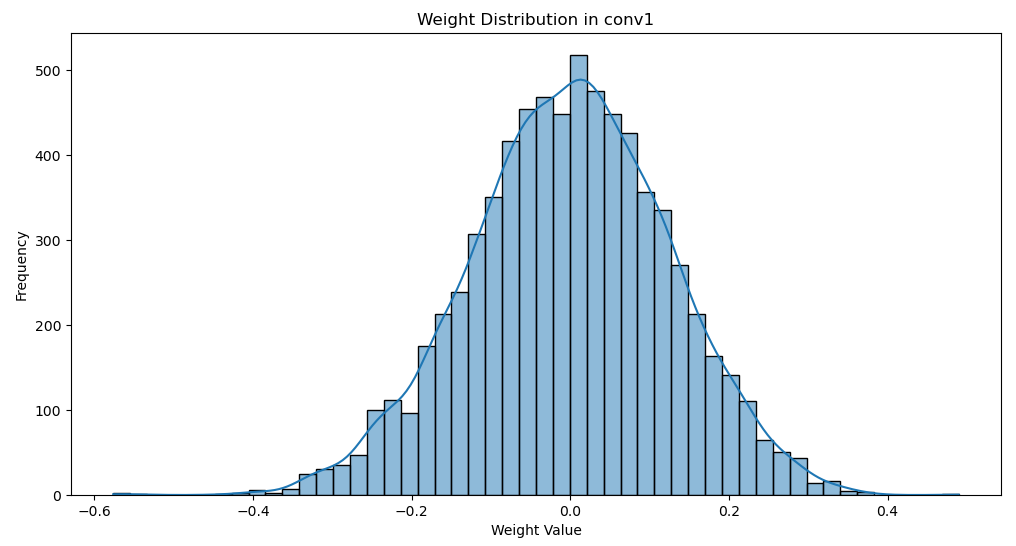}
  \centering
  \textbf{(B) Backpropagation}
\end{minipage}
}
\caption{Weight distribution of \textbf{Optimal-HardWTA} and \textbf{Backpropagation} configurations for CIFAR-10 at Layer 1.}
\label{fig:weights-dist2}
\end{figure}

Analysis of weight distributions (Figures \ref{fig:weights-dist1} and \ref{fig:weights-dist2}) revealed distinct patterns across learning approaches. \path{SoftWTA} (Figure \ref{fig:weights-dist1}A) maintained a broad normal distribution across all layers, reflecting its soft competition approach. In contrast, \path{Lagani-HardWTA} (Figure \ref{fig:weights-dist1}B) showed a bimodal distribution with two distinct peaks, suggesting more binary weight patterns emerging from the hard competition mechanism.

Our \path{Optimal-HardWTA}'s BCM learning rule (Figure \ref{fig:weights-dist2}A) produced a highly sparse distribution with a pronounced peak near zero and long tails, closely mirroring the sparse connectivity observed in biological neural networks \citep{hawkins2016biological}. \path{Backpropagation} (Figure \ref{fig:weights-dist2}B) displayed a classic normal distribution, characteristic of the global nature of gradient descent optimization.

\subsubsection{PGA Receptive Fields}

\begin{figure}[h!]
  \centering
  \resizebox{0.99\textwidth}{!}{%
    \begin{minipage}[b]{0.32\textwidth}
      \includegraphics[width=\textwidth]{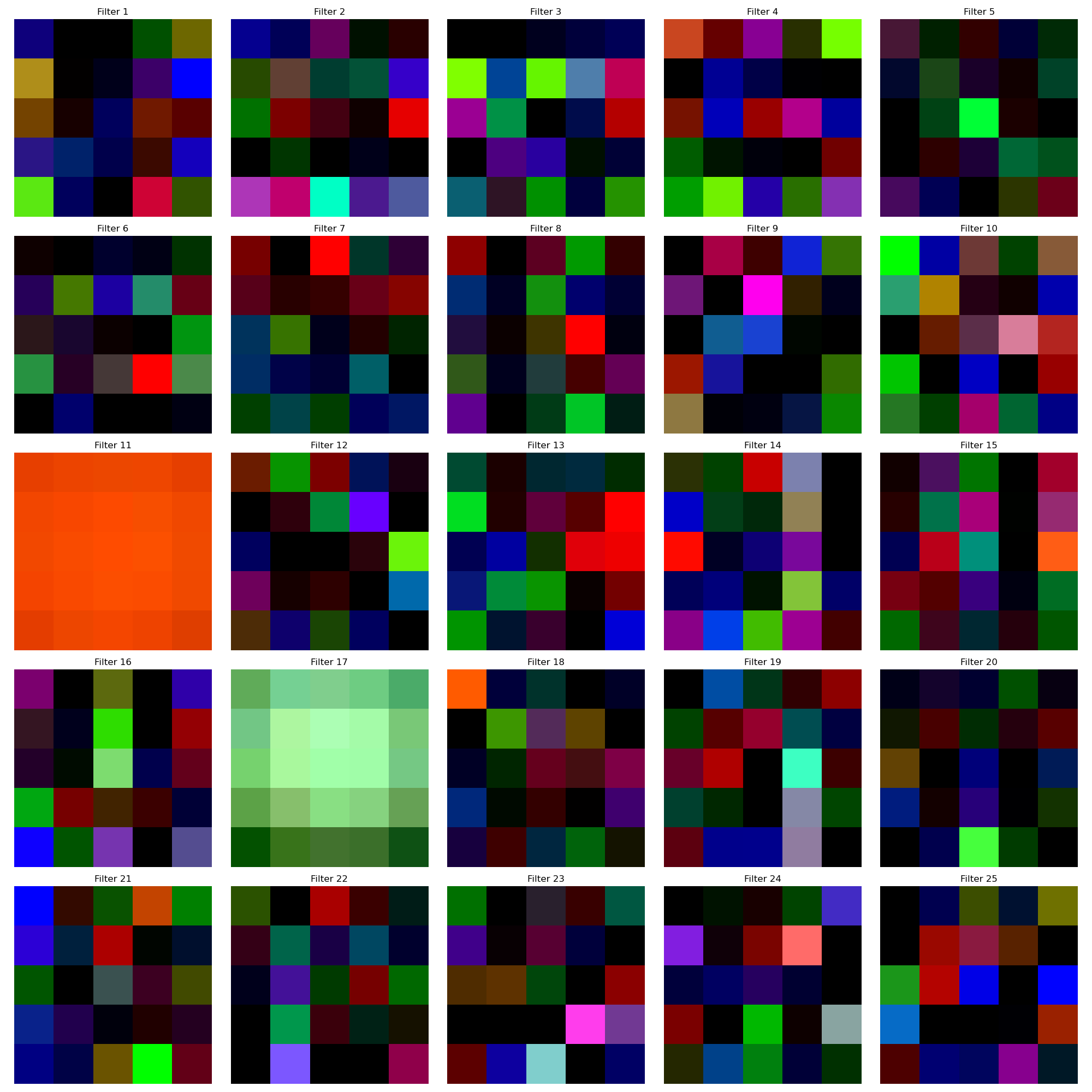}
      \centering
      \textbf{(A) Layer 1}
    \end{minipage}
    \hfill
    \begin{minipage}[b]{0.32\textwidth}
      \includegraphics[width=\textwidth]{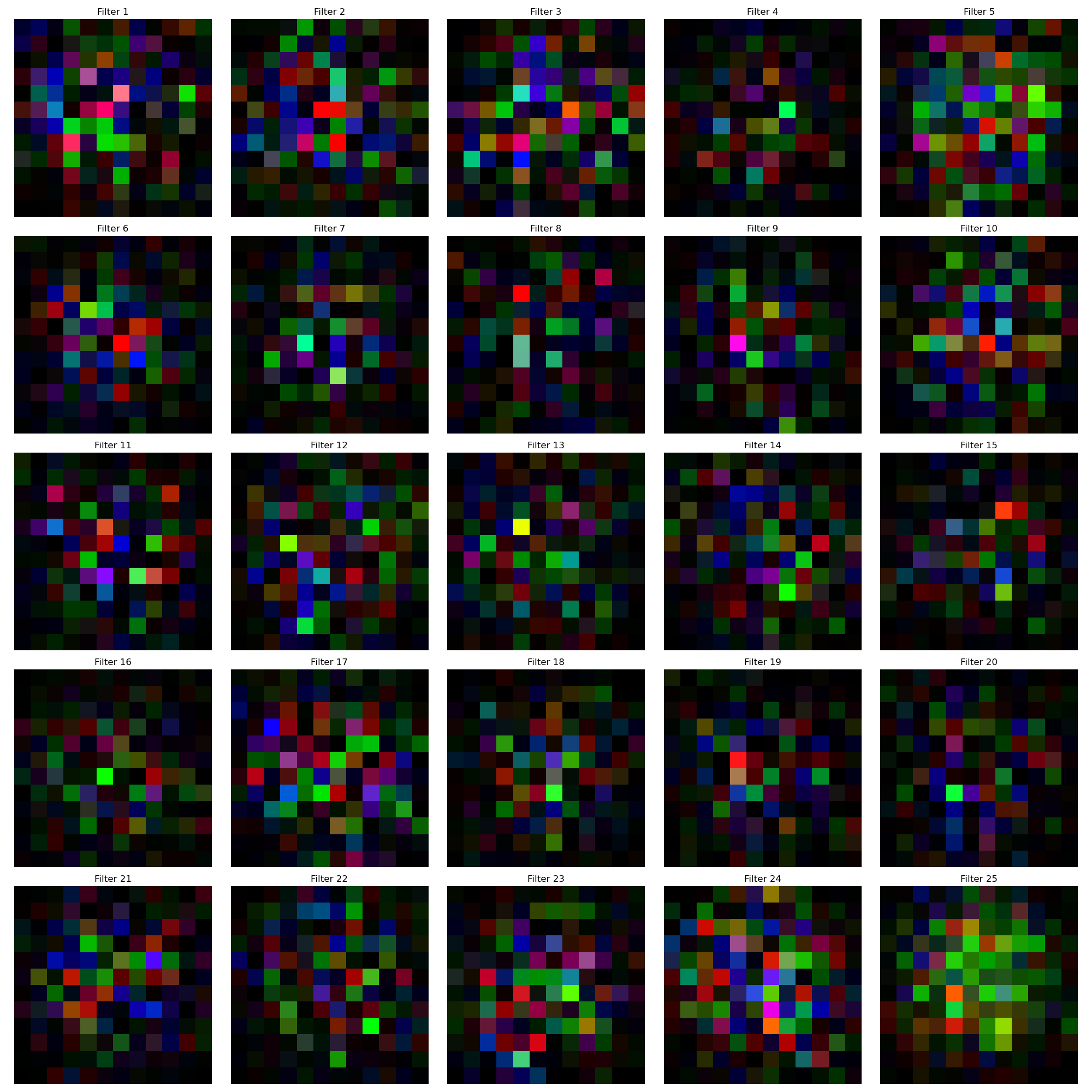}
      \centering
      \textbf{(B) Layer 2}
    \end{minipage}
    \begin{minipage}[b]{0.32\textwidth}
      \includegraphics[width=\textwidth]{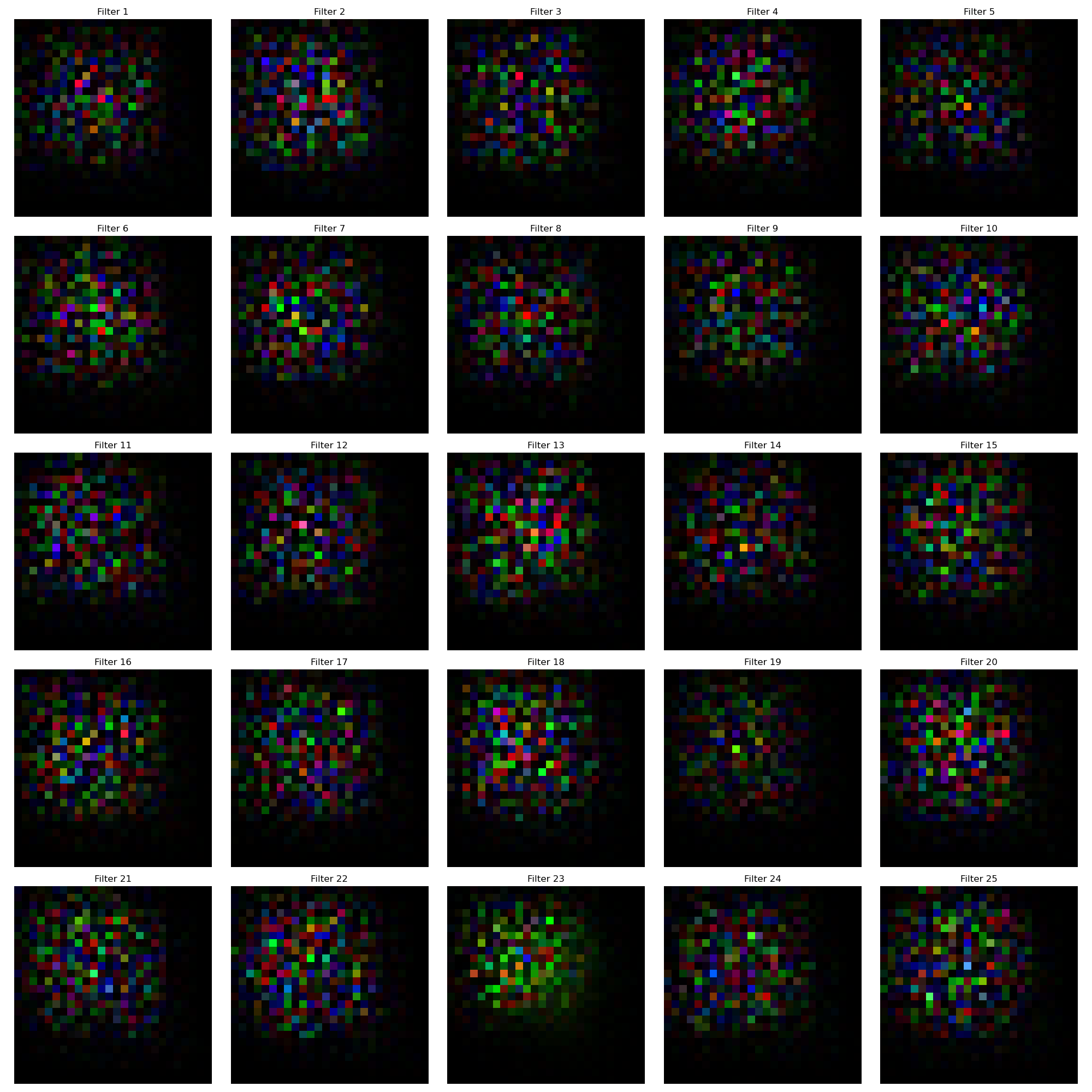}
      \centering
      \textbf{(C) Layer 3}
    \end{minipage}
  }
  \caption{PGA Receptive fields for first 25 neurons at each layer of the \textbf{SoftWTA} configuration for CIFAR-10.}
  \label{fig:rf-softhebb}
\end{figure}

\textbf{SoftWTA} (Figure \ref{fig:rf-softhebb}) demonstrated features that deviate from traditional Gabor-like patterns. Layer 1 exhibited diverse patterns including solid colour detectors (visible in bright orange/green squares), chequerboard patterns capturing local contrast and various colour-mixing effects. Layer 2 showed a transition to more complex combinations with fragmented patterns, smaller pixel clusters, and intricate colour interactions. Layer 3 developed highly abstract, sparse representations with noise-like features and scattered colour points, suggesting specialised detectors for complex distributed patterns.

\begin{figure}[h!]
  \centering
  \resizebox{0.99\textwidth}{!}{%
    \begin{minipage}[b]{0.32\textwidth}
      \includegraphics[width=\textwidth]{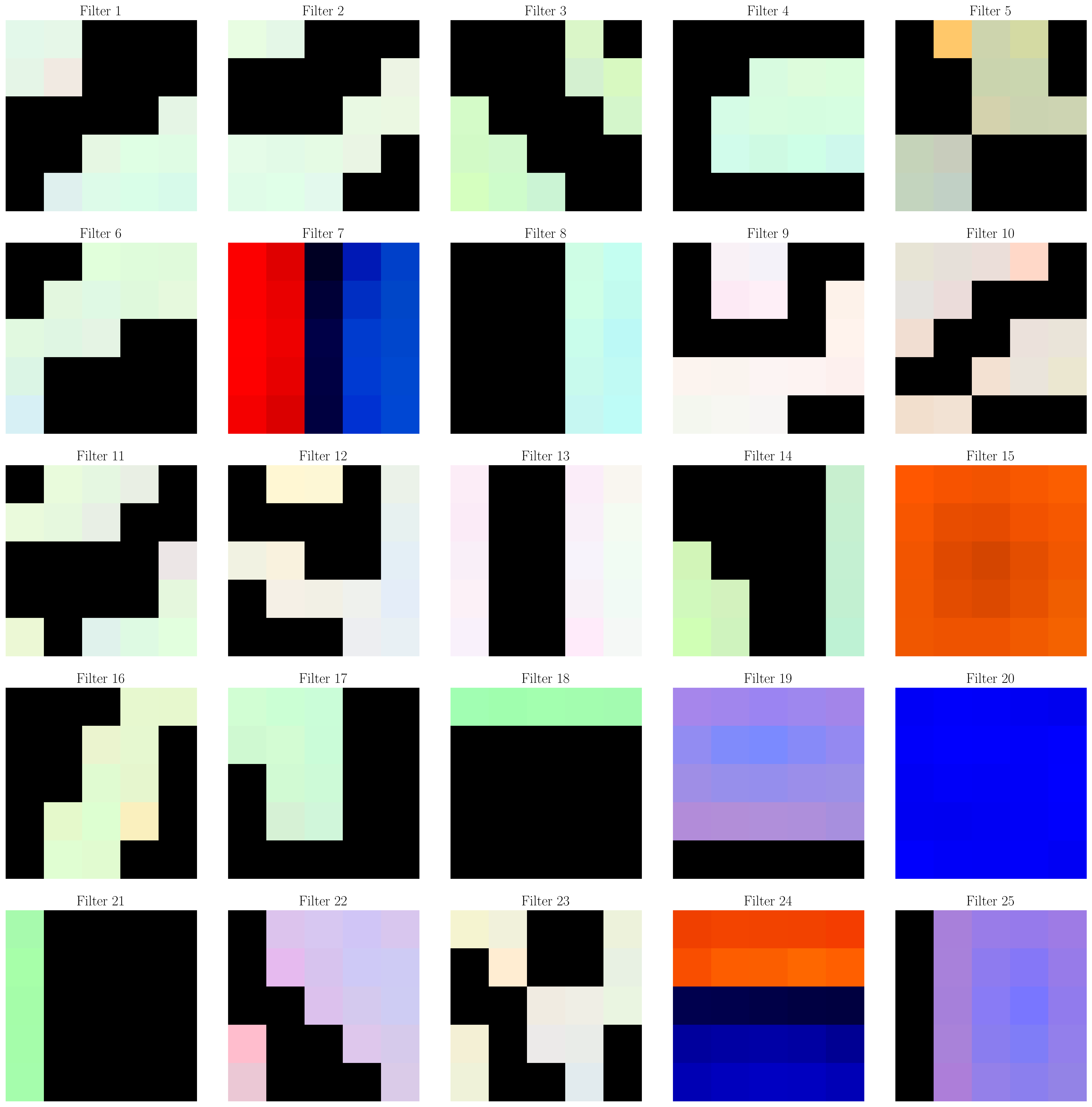}
      \centering
      \textbf{(A) Layer 1}
    \end{minipage}
    \hfill
    \begin{minipage}[b]{0.32\textwidth}
      \includegraphics[width=\textwidth]{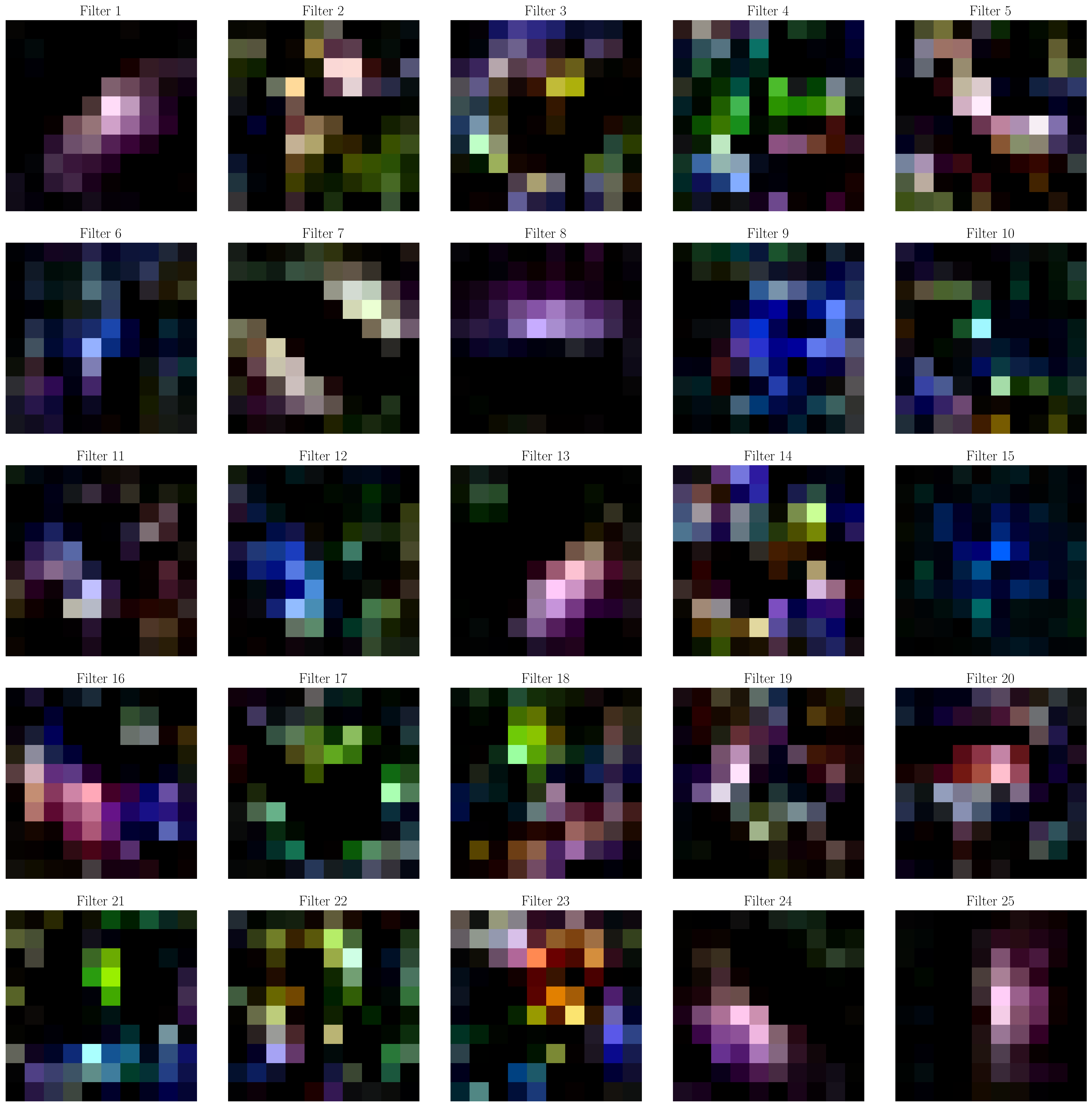}
      \centering
      \textbf{(B) Layer 2}
    \end{minipage}
    \begin{minipage}[b]{0.32\textwidth}
      \includegraphics[width=\textwidth]{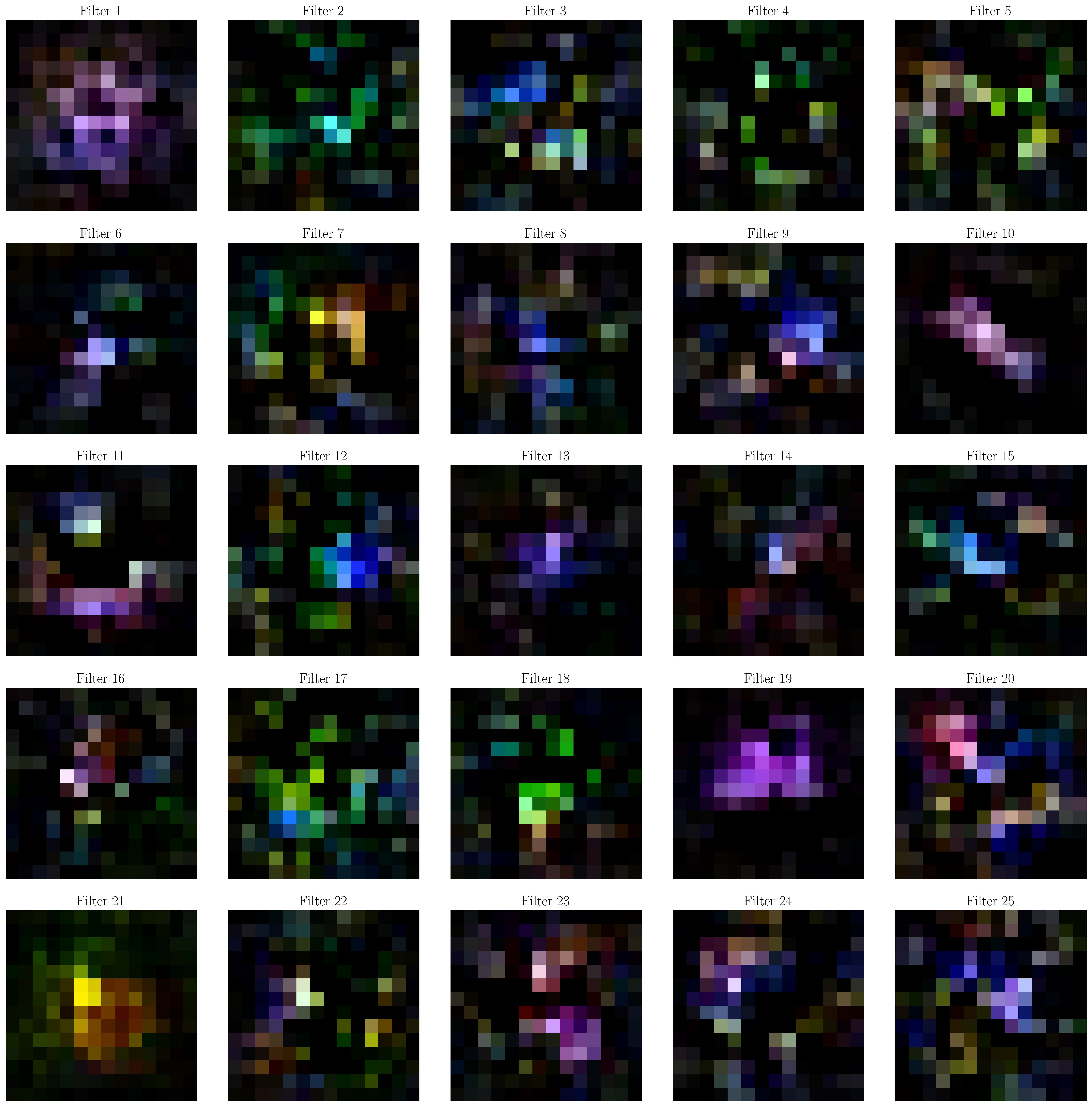}
      \centering
      \textbf{(C) Layer 3}
    \end{minipage}
  }
  \caption{PGA Receptive fields for first 25 neurons at each layer of the \textbf{Lagani-HardWTA} configuration for CIFAR-10.}
  \label{fig:rf-lagani}
\end{figure}

\textbf{Lagani-HardWTA} (Figure \ref{fig:rf-lagani}) exhibited classical biological neural network characteristics with Gabor-like edge detection. Layer 1 showed strong binary patterns with sharp edges and clear geometric shapes, where black-and-white regions and isolated pure colours (purple, blue, orange) demonstrate decisive edge detection from hard-WTA competition. Layer 2 maintained structured organisation with discernible orientation shapes and subtle colour blending, indicating effective combination of edge detectors with colour information. Layer 3 preserved defined shapes compared to \path{SoftWTA}'s diffuse activations, reflecting maintained feature selectivity in deeper layers.

\begin{figure}[h!]
  \centering
  \resizebox{0.99\textwidth}{!}{%
    \begin{minipage}[b]{0.32\textwidth}
      \includegraphics[width=\textwidth]{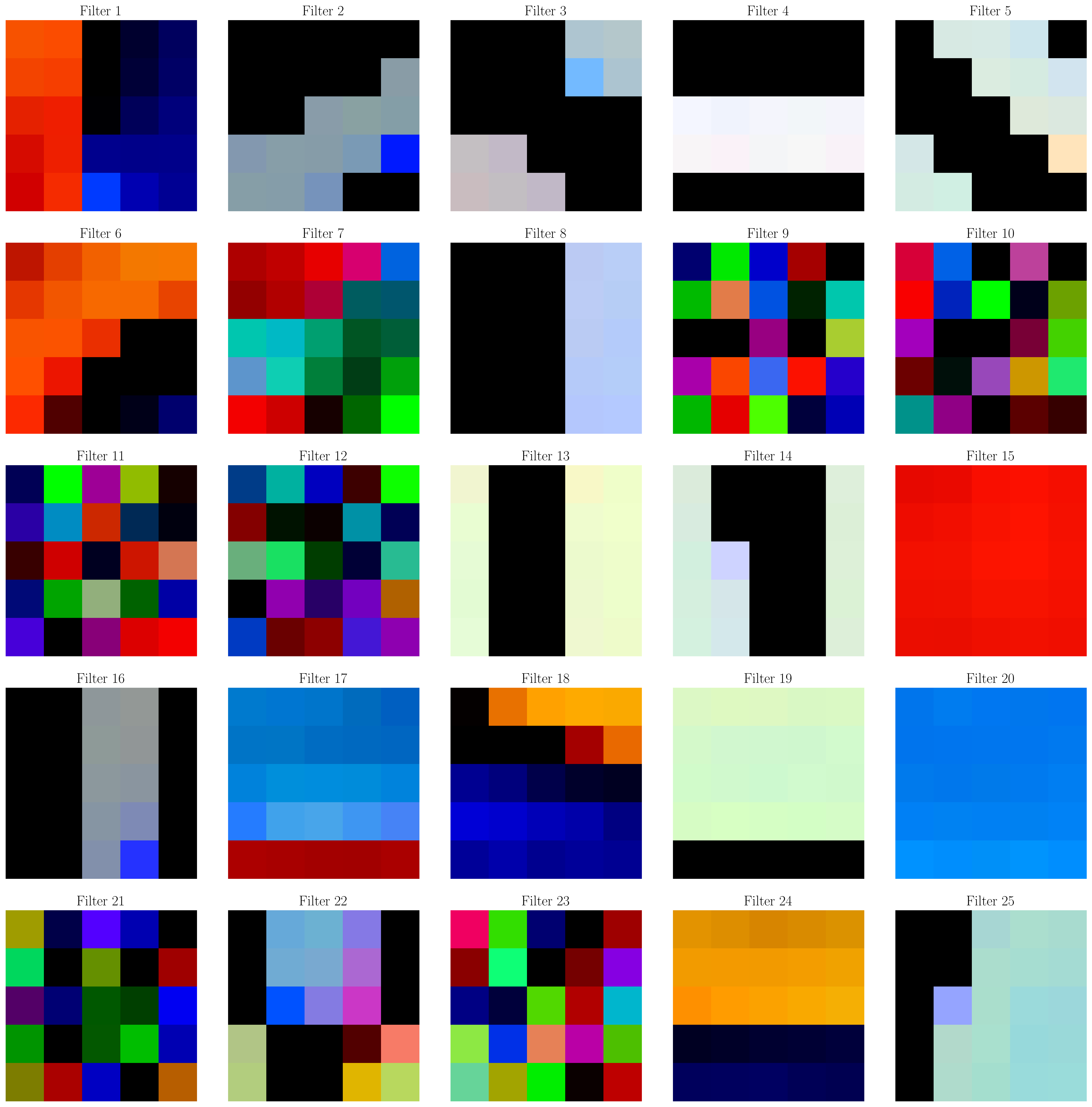}
      \centering
      \textbf{(A) Layer 1}
    \end{minipage}
    \hfill
    \begin{minipage}[b]{0.32\textwidth}
      \includegraphics[width=\textwidth]{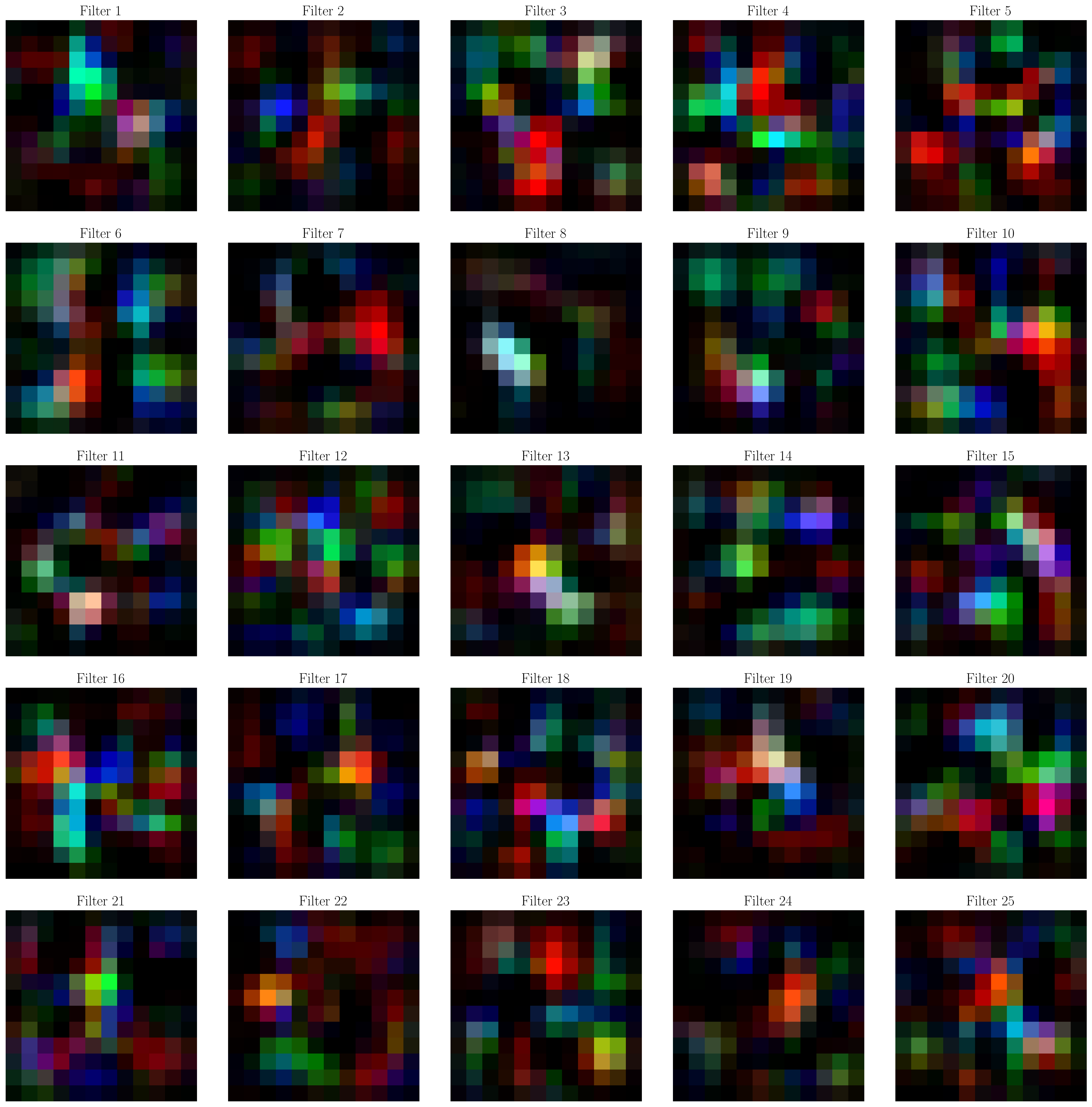}
      \centering
      \textbf{(B) Layer 2}
    \end{minipage}
    \begin{minipage}[b]{0.32\textwidth}
      \includegraphics[width=\textwidth]{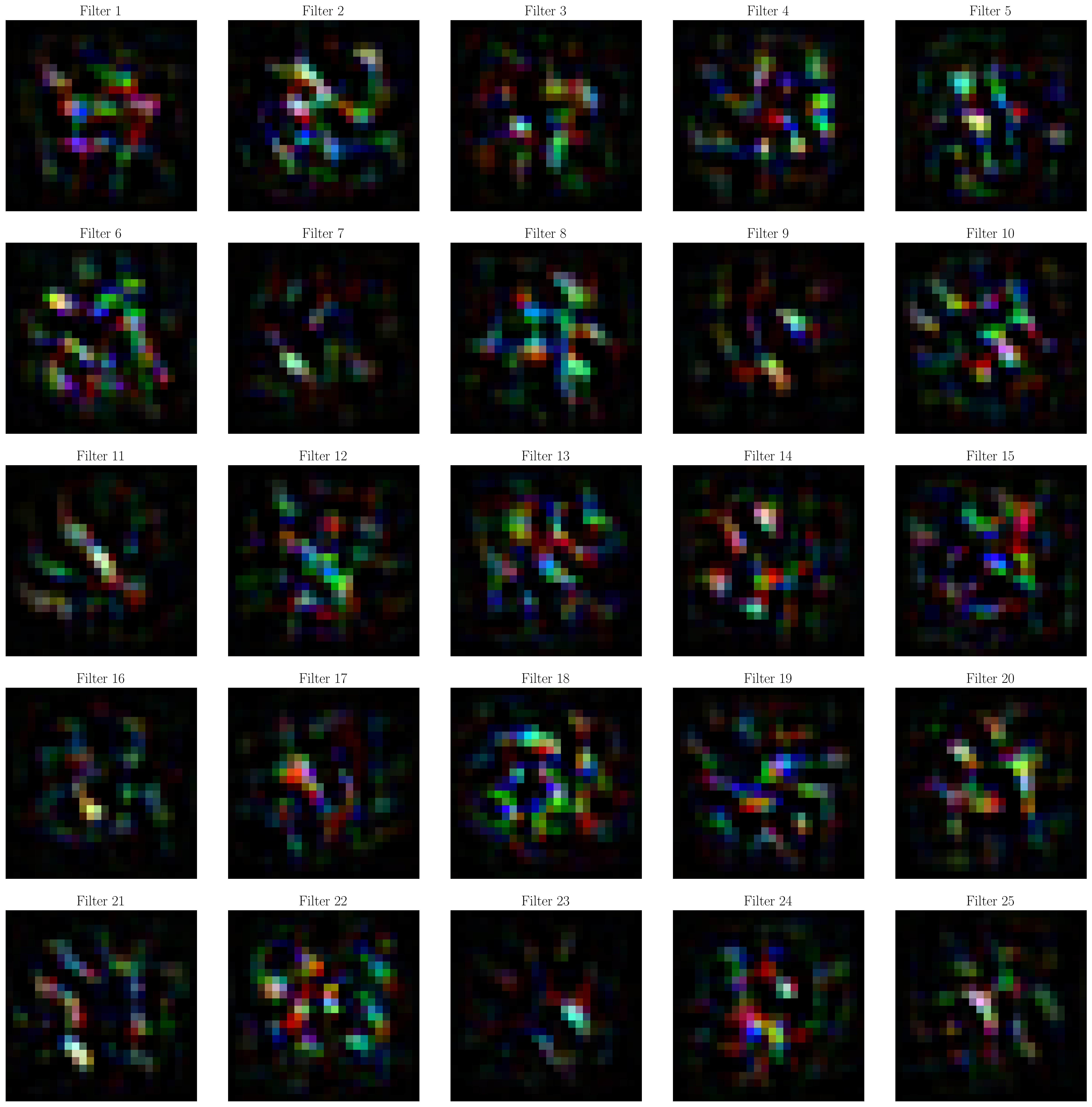}
      \centering
      \textbf{(C) Layer 3}
    \end{minipage}
  }
  \caption{PGA Receptive fields for first 25 neurons at each layer of the \textbf{Optimal-HardWTA} configuration for CIFAR-10.}
  \label{fig:rf-best}
\end{figure}

\begin{figure}[h!]
  \centering
  \resizebox{0.99\textwidth}{!}{%
    \begin{minipage}[b]{0.32\textwidth}
      \includegraphics[width=\textwidth]{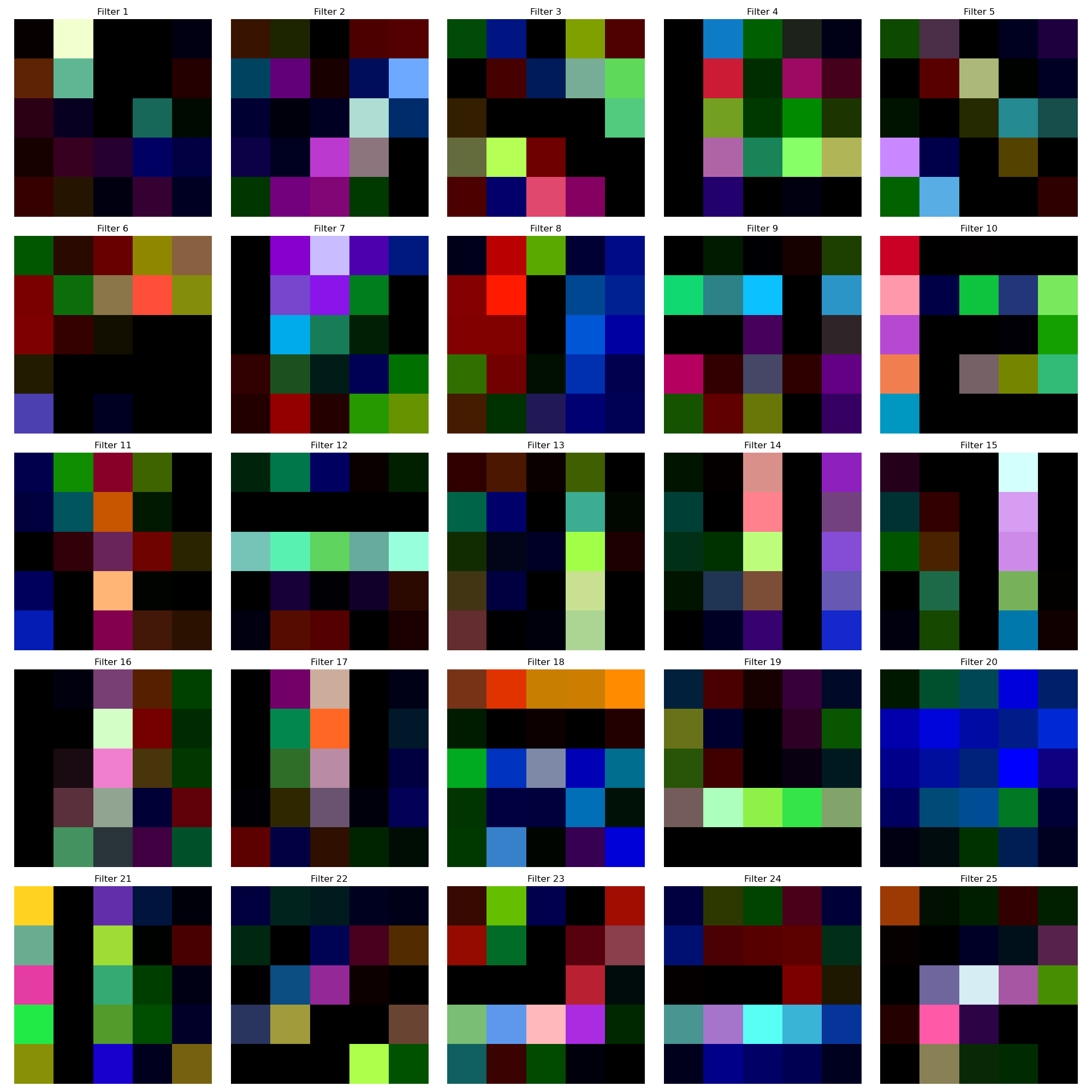}
      \centering
      \textbf{(A) Layer 1}
    \end{minipage}
    \hfill
    \begin{minipage}[b]{0.32\textwidth}
      \includegraphics[width=\textwidth]{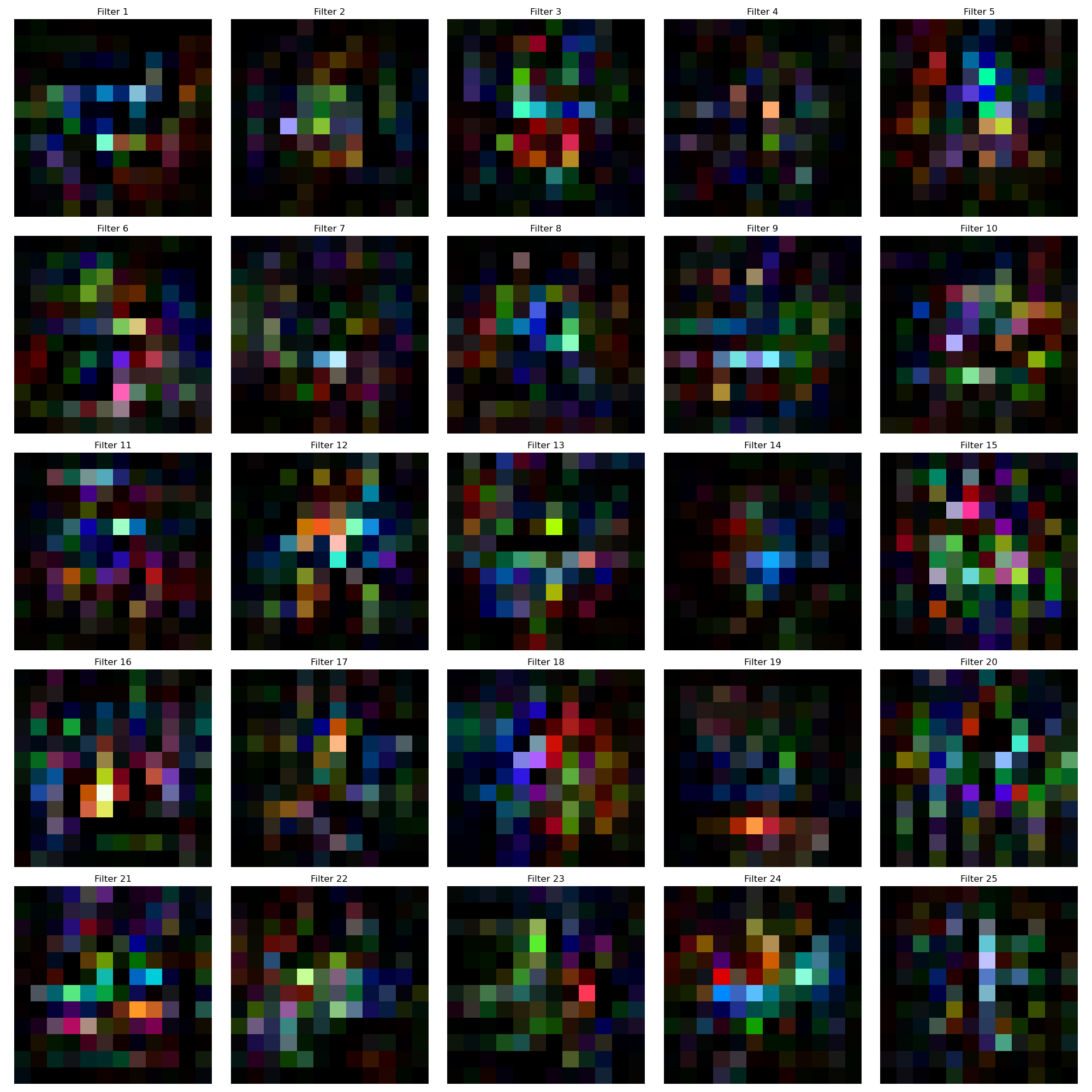}
      \centering
      \textbf{(B) Layer 2}
    \end{minipage}
    \begin{minipage}[b]{0.32\textwidth}
      \includegraphics[width=\textwidth]{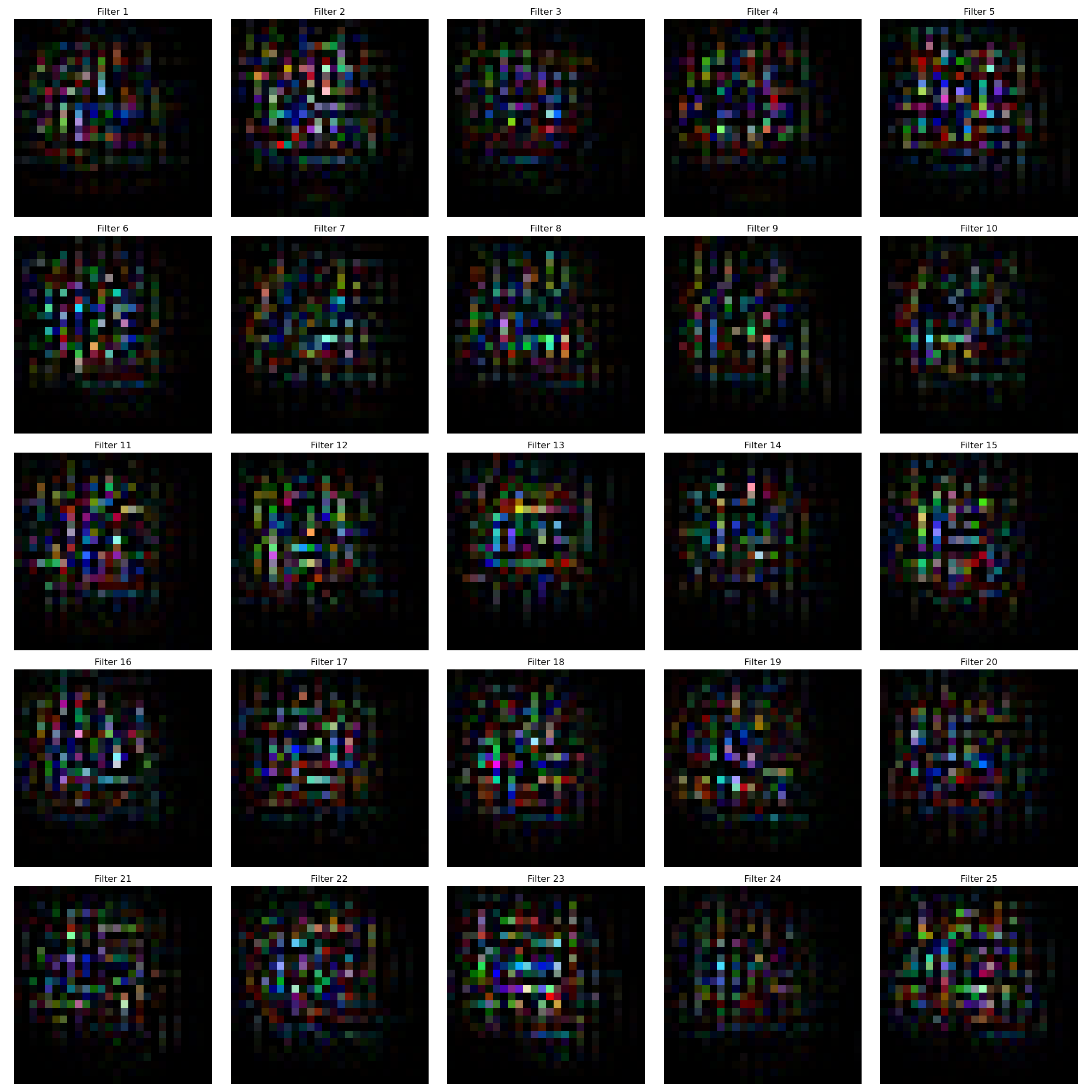}
      \centering
      \textbf{(C) Layer 3}
    \end{minipage}
  }
  \caption{PGA Receptive fields for first 25 neurons at each layer of the \textbf{Backpropagation} configuration for CIFAR-10.}
  \label{fig:rf-Backpropagation}
\end{figure}

\textbf{Optimal-HardWTA} (Figure \ref{fig:rf-best}) showcased successful integration of hard-WTA, lateral inhibition and BCM learning. Layer 1 combined sharp geometric boundaries comparable to Lagani's Gabor-like edge detection, features that characterises biological neural networks, with more sophisticated colour processing, demonstrated through solid or multiple-colour colour detectors and complex edge patterns with smoother gradients and colour information. Layer 2 exhibited refined feature organisation with distinct colour combinations and organised spatial arrangements, where lateral inhibition promoted coherent transitions between spatial regions. Layer 3 developed specialised detectors with sparse but organised patterns maintaining structural coherence, suggesting effective preservation of feature selectivity while enabling complex pattern detection.

\textbf{Backpropagation} (Figure \ref{fig:rf-Backpropagation}) revealed characteristics of gradient-based optimisation. Layer 1 demonstrated strong directional selectivity with vertical and horizontal structures alongside systematic colour organisation, appearing more regularly than in Hebbian models. Layer 2 developed complex features while maintaining organisational principles, with uniform pattern and colour distribution suggesting global optimisation benefits. Layer 3 exhibited distinctive grid-like patterns with architectural regularity, reflecting systematic feature decomposition optimised for classification rather than biological plausibility.

\subsection{Limitations}


Several limitations were also encountered. While competitive with backpropagation in shallow networks, Hebbian learning may face challenges in deeper architectures, as the width-factor significantly increases the parameters in the architecture by a factor of 16x with each additional layer. Enhanced feature representation by eliminating redundant information among neurons might help resolve this problem. The small pixel images and 3-4 layer models limited the complexity of learnt representations, particularly in deeper layers. Performance varied significantly with architectural changes, highlighting the need for careful design considerations. Furthermore, the study focused on complete clean datasets, and limited training data scenarios and adversarial attacks remain to be explored.

\section{Discussion}\label{discuss}

This work advances the state-of-the-art in biologically-plausible Convolutional Neural Networks by demonstrating that Hard Winner-Takes-All (WTA) competition can achieve performance comparable to backpropagation while maintaining biological realism.  By reaching 75.2\% mean accuracy on CIFAR-10, our implementation significantly improves upon previous Hard-WTA Hebbian approaches of the same network depth (64.6\% by \cite{miconi2017biologically}) and approximates backpropagation performance (mean 75.2\%), albeit only in shallow networks, through careful integration of lateral inhibition and BCM learning rule. These results extend across datasets, achieving backpropagation-level performance on MNIST (98\%) and superior on STL-10 (69.5\% mean in under 20 epochs), all while demonstrating exceptional consistency across experimental runs.

Our results revealed that the performance of the Optimal-Hard WTA configuration was significantly superior to that of the Lagani-Hard WTA across all three datasets during the latter half of the testing period and at the end of it. Furthermore, it closely matched the accuracy achieved by Backpropagation and Soft WTA, outperforming both methodologies with the STL-10 under restricted train-time scenarios. These findings highlight the robustness and generality of the superior performance within the scope of this research.

The superiority of Hard-WTA over Soft-WTA \citep{journe2022hebbian} stems from its modelling of biological neural dynamics, particularly sparse coding where only a small fraction of neurons are active for any input. This sparsity translates into practical benefits: improved feature interpretability and reduced computational overhead during both training and inference.

Our approach addresses fundamental limitations of backpropagation by eliminating the need for labelled data and global error signals. The local nature of Hebbian learning, combined with Hard-WTA competition, enables efficient parallel processing and reduced memory requirements, critical advantages for neuromorphic computing and edge applications. Moreover, the emergence of orientation-selective neurons and centre-surround receptive fields in our network provides computational neuroscientists with a more faithful model of biological visual processing.

Two key challenges remain: scaling to larger datasets and enhancing biological realism. Future work should focus on developing efficient learning schemes for high-resolution images while controlling parameter growth in deeper architectures. Additionally, incorporating distinct excitatory and inhibitory populations could further bridge the gap between artificial and biological networks while potentially improving representation learning.

This research demonstrates that embracing biological constraints, particularly Hard-WTA competition and local Hebbian learning rules, can enhance both the efficiency and robustness of artificial neural networks while maintaining competitive accuracy. Our framework validates that principles underlying biological neural computation can be successfully translated into practical artificial systems, opening new possibilities for energy-efficient, interpretable, and robust AI architectures that better reflect biological intelligence.

\section*{Data and Code Availability}

All code used in this study are publicly available. The code is available on GitHub: \url{https://github.com/Julian-JN/Advancing-the-Biological-Plausibility-and-Efficacy-of-Hebbian-Convolutional-Neural-Networks}. Any requests for additional data or materials should be directed to the corresponding author.

\section*{Declaration of competing interest}

No competing interests

\bibliographystyle{elsarticle-harv} 
\bibliography{bibliography.bib}

\newpage
\appendix
\setcounter{section}{0}  
\renewcommand{\thesection}{Appendix \Alph{section}}  
\renewcommand{\thesubsection}{\Alph{section}.\arabic{subsection}}  

\section{Details of Methods}

\subsection{Derivation of Equation 8 from Equation 2} \label{derivation}

Equation 2 defines the Grossberg Instar rule:

\begin{equation}
\Delta w = \eta y (x - w) = \eta (yx - yw).
\end{equation}

where: \( y \) is the post-synaptic activation, \( x \) is the pre-synaptic input, \( w \) is the synaptic weight, \( yx \) is the Hebbian learning term and \( yw \) is a normalisation term preventing unbounded growth.

In a convolutional neural network (CNN), weight updates occur over localised receptive fields rather than single synapses. To extend this to CNNs, we replace the simple product \( yx \) with a convolution operation $*$:

\begin{equation}
yx = x * y,
\end{equation}

where \( * \) denotes a 2D convolution:

\begin{equation}
(y * x)_{c, h, w} = \sum_{i,j} x_{c, h+i, w+j} y_{c, i, j}.
\end{equation}

Similarly, the second term \( yw \), which in the original rule represents a decay proportional to \( y \), is implemented as:

\begin{equation}
yw = (\sum_{b,h,w} y_{b,h,w}) w.
\end{equation}

Thus, the final convolutional update rule is:

\begin{equation}
\Delta w = (y * x) - yw.
\end{equation}

This ensures weight updates occur based on local convolutional correlations while maintaining stability across spatial locations. This functions as a GPU optimisation of \cite{amato2019hebbian}'s research and also incorporates the Grossberg derivation applied by \cite{journe2022hebbian}.

\subsection{SoftWTA Weight Initialisation} \label{weight init}

The weight initialisation scheme used for the SoftWTA Hebbian (\path{SoftWTA}) configuration is detailed below, mirroring the Normal Distribution used in \citep{journe2022hebbian}. To allow all neurons to learn and reach convergence in a single epoch, a positive distribution with a radius larger than 2.5 is necessary:

\begin{equation*}
\begin{aligned}
\text{weight\_range} &= \frac{25}{\sqrt{C_{in} \cdot K_h \cdot K_w}} \\[10pt]
W &= \text{weight\_range} \cdot X \\[5pt]
X &\sim \mathcal{N}(0, 1) \\[10pt]
\end{aligned}
\end{equation*}

\section{Centre-Surround Filters through Dale's Principle} \label{centre-surr}

\begin{figure}[h!]
  \centering
  \resizebox{0.95\textwidth}{!}{%
    \begin{minipage}[b]{0.499\textwidth}
      \includegraphics[width=\textwidth]{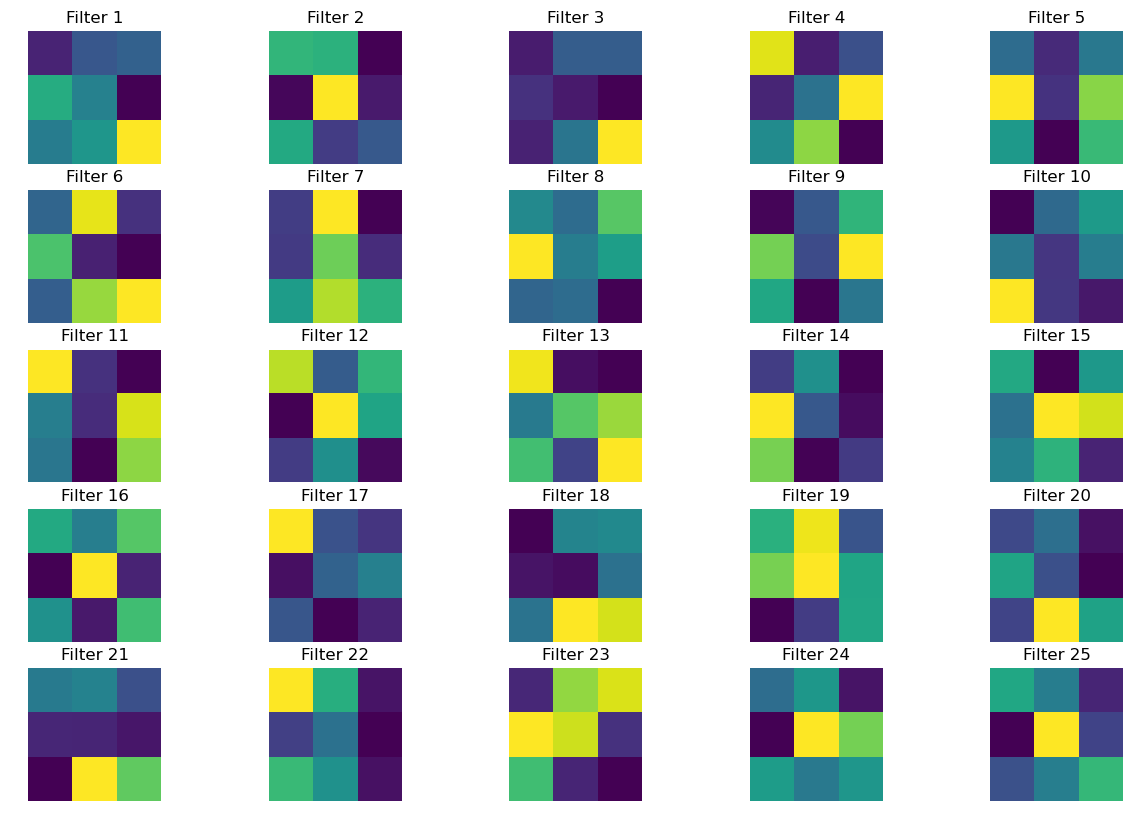}
      \centering
      \textbf{(A) Initial Layer 2}
    \end{minipage}
    \hfill
    \begin{minipage}[b]{0.499\textwidth}
      \includegraphics[width=\textwidth]{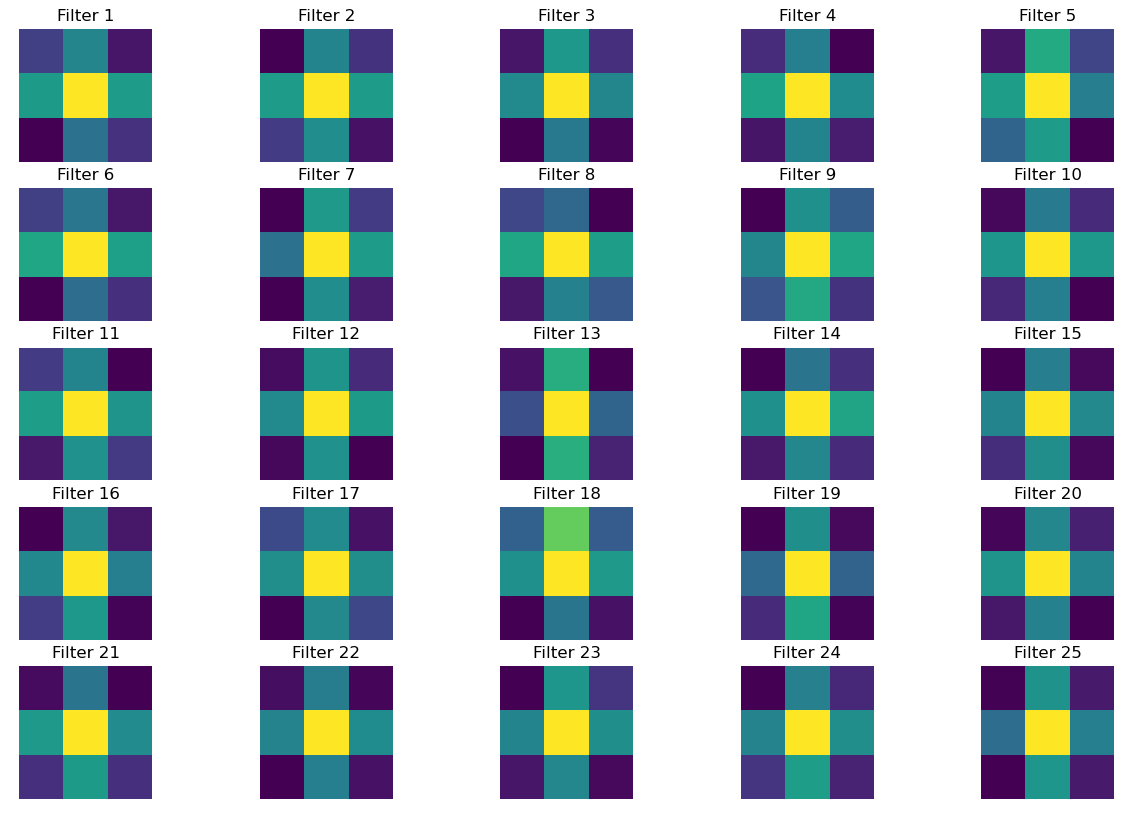}
      \centering
      \textbf{(B) Trained Layer 2}
    \end{minipage}
  }
  \caption{Direct weight visualisation of filters in Layer 2 of \textbf{Dale-Depthwise-Surr/HardWTA-BCM} before and after training. Note the exclusive formation of centre-surround filters.}
  \label{fig:dale-direct}
\end{figure}

The depthwise network (\path{Dale_Depthwise-Surr/HardWTA/Cos-BCM}) which fully respected Dale's principle exhibited exclusively centre-surround patterns in its depthwise filters at Layers 2-3 (seen through the direct filter visualisation in Figure \ref{fig:dale-direct}), further accentuating the biological realism of combining the depthwise architecture with Hebbian Learning and its biologically-inspired competition mechanisms.

\section{Network Architectures} \label{architectures}


\begin{table}[htbp]
    \centering
    \caption{Journe CNN Architecture (Non-Depthwise)}
    \label{tab:softhebb-architecture-non-depthwise}
    \resizebox{\textwidth}{!}{%
    \setlength{\tabcolsep}{6pt}
    \renewcommand{\arraystretch}{1.2}
    \begin{tabular}{|l|l|l|c|c|c|l|}
    \hline
    \textbf{Layer} & \textbf{Type} & \textbf{Output Shape} & \textbf{Kernel} & \textbf{Stride} & \textbf{Padding} & \textbf{Activation} \\
    \hline
    Input & -- & (3, 32, 32) & -- & -- & -- & -- \\
    \hline
    1 & BatchNorm2d & (3, 32, 32) & -- & -- & -- & -- \\
     & HebbianConv2d & (96, 32, 32) & 5x5 & 1 & 2 & -- \\
     & Triangle & (96, 32, 32) & -- & -- & -- & power=0.7 \\
     & MaxPool2d & (96, 16, 16) & 4x4 & 2 & 1 & -- \\
    \hline
    2 & BatchNorm2d & (96, 16, 16) & -- & -- & -- & -- \\
     & HebbianConv2d & (384, 16, 16) & 3x3 & 1 & 1 & -- \\
     & Triangle & (384, 16, 16) & -- & -- & -- & power=1.4 \\
     & MaxPool2d & (384, 8, 8) & 4x4 & 2 & 1 & -- \\
    \hline
    3 & BatchNorm2d & (384, 8, 8) & -- & -- & -- & -- \\
     & HebbianConv2d & (1536, 8, 8) & 3x3 & 1 & 1 & -- \\
     & Triangle & (1536, 8, 8) & -- & -- & -- & power=1.0 \\
     & AvgPool2d & (1536, 4, 4) & 2x2 & 2 & 0 & -- \\
    \hline
    Output & Linear & (10) & -- & -- & -- & -- \\
    \hline
    \end{tabular}%
    }
\end{table}

\begin{table}[htbp]
    \centering
    \caption{Lagani 3-layer CNN Architecture (Non-Depthwise)}
    \label{tab:lagani-short-architecture-non-depthwise}
    \resizebox{\textwidth}{!}{%
    \setlength{\tabcolsep}{6pt}
    \renewcommand{\arraystretch}{1.2}
    \begin{tabular}{|l|l|l|c|c|c|l|}
    \hline
    \textbf{Layer} & \textbf{Type} & \textbf{Output Shape} & \textbf{Kernel} & \textbf{Stride} & \textbf{Padding} & \textbf{Activation} \\
    \hline
    Input & -- & (3, 32, 32) & -- & -- & -- & -- \\
    \hline
    1 & BatchNorm2d & (3, 32, 32) & -- & -- & -- & -- \\
     & HebbianConv2d & (96, 28, 28) & 5x5 & 1 & 0 & Cosine \\
     & Triangle & (96, 28, 28) & -- & -- & -- & power=1.0 \\
     & MaxPool2d & (96, 14, 14) & 2x2 & 2 & 0 & -- \\
    \hline
    2 & BatchNorm2d & (96, 14, 14) & -- & -- & -- & -- \\
     & HebbianConv2d & (128, 12, 12) & 3x3 & 1 & 0 & Cosine \\
     & Triangle & (128, 12, 12) & -- & -- & -- & power=1.0 \\
    \hline
    3 & BatchNorm2d & (128, 12, 12) & -- & -- & -- & -- \\
     & HebbianConv2d & (192, 10, 10) & 3x3 & 1 & 0 & Cosine \\
     & Triangle & (192, 10, 10) & -- & -- & -- & power=1.0 \\
     & AvgPool2d & (192, 5, 5) & 2x2 & 2 & 0 & -- \\
    \hline
    Output & Linear & (10) & -- & -- & -- & -- \\
    \hline
    \end{tabular}%
    }
\end{table}

\begin{table}[htbp]
    \centering
    \caption{Lagani 4-layer CNN Architecture (Non-Depthwise)}
    \label{tab:lagani-long-architecture-non-depthwise}
    \resizebox{\textwidth}{!}{%
    \setlength{\tabcolsep}{6pt}
    \renewcommand{\arraystretch}{1.2}
    \begin{tabular}{|l|l|l|c|c|c|l|}
    \hline
    \textbf{Layer} & \textbf{Type} & \textbf{Output Shape} & \textbf{Kernel} & \textbf{Stride} & \textbf{Padding} & \textbf{Activation} \\
    \hline
    Input & -- & (3, 32, 32) & -- & -- & -- & -- \\
    \hline
    1 & BatchNorm2d & (3, 32, 32) & -- & -- & -- & -- \\
     & HebbianConv2d & (96, 28, 28) & 5x5 & 1 & 0 & Cosine \\
     & Triangle & (96, 28, 28) & -- & -- & -- & power=1.0 \\
     & MaxPool2d & (96, 14, 14) & 2x2 & 2 & 0 & -- \\
    \hline
    2 & BatchNorm2d & (96, 14, 14) & -- & -- & -- & -- \\
     & HebbianConv2d & (128, 12, 12) & 3x3 & 1 & 0 & Cosine \\
     & Triangle & (128, 12, 12) & -- & -- & -- & power=1.0 \\
    \hline
    3 & BatchNorm2d & (128, 12, 12) & -- & -- & -- & -- \\
     & HebbianConv2d & (192, 10, 10) & 3x3 & 1 & 0 & Cosine \\
     & Triangle & (192, 10, 10) & -- & -- & -- & power=1.0 \\
     & AvgPool2d & (192, 5, 5) & 2x2 & 2 & 0 & -- \\
    \hline
    4 & BatchNorm2d & (192, 5, 5) & -- & -- & -- & -- \\
     & HebbianConv2d & (256, 3, 3) & 3x3 & 1 & 0 & Cosine \\
     & Triangle & (256, 3, 3) & -- & -- & -- & power=1.0 \\
    \hline
    Output & Linear & (10) & -- & -- & -- & -- \\
    \hline
    \end{tabular}%
    }
\end{table}

\begin{table}[htbp]
    \centering
    \caption{Journe 4-layer CNN Architecture (Non-Depthwise)}
    \label{tab:softhebb-architecture-non-depthwise}
    \resizebox{\textwidth}{!}{%
    \setlength{\tabcolsep}{6pt}
    \renewcommand{\arraystretch}{1.2}
    \begin{tabular}{|l|l|l|c|c|c|l|}
    \hline
    \textbf{Layer} & \textbf{Type} & \textbf{Output Shape} & \textbf{Kernel} & \textbf{Stride} & \textbf{Padding} & \textbf{Activation} \\
    \hline
    Input & -- & (3, 32, 32) & -- & -- & -- & -- \\
    \hline
    1 & BatchNorm2d & (3, 32, 32) & -- & -- & -- & -- \\
     & HebbianConv2d & (96, 32, 32) & 5x5 & 1 & 2 & -- \\
     & Triangle & (96, 32, 32) & -- & -- & -- & power=0.7 \\
     & MaxPool2d & (96, 16, 16) & 4x4 & 2 & 1 & -- \\
    \hline
    2 & BatchNorm2d & (96, 16, 16) & -- & -- & -- & -- \\
     & HebbianConv2d & (384, 16, 16) & 3x3 & 1 & 1 & -- \\
     & Triangle & (384, 16, 16) & -- & -- & -- & power=1.4 \\
     & MaxPool2d & (384, 8, 8) & 4x4 & 2 & 1 & -- \\
    \hline
    3 & BatchNorm2d & (384, 8, 8) & -- & -- & -- & -- \\
     & HebbianConv2d & (1536, 8, 8) & 3x3 & 1 & 1 & -- \\
     & Triangle & (1536, 8, 8) & -- & -- & -- & power=1.0 \\
     & AvgPool2d & (1536, 4, 4) & 2x2 & 2 & 0 & -- \\
    \hline
    4 & BatchNorm2d & (1536, 4, 4) & -- & -- & -- & -- \\
     & HebbianConv2d & (6144, 4, 4) & 3x3 & 1 & 1 & -- \\
     & Triangle & (6144, 4, 4) & -- & -- & -- & power=1.0 \\
     & AvgPool2d & (6144, 2, 2) & 2x2 & 2 & 0 & -- \\
    \hline
    Output & Linear & (10) & -- & -- & -- & -- \\
    \hline
    \end{tabular}%
    }
\end{table}

\begin{table}[htbp]
    \centering
    \caption{Journe CNN Architecture (Depthwise)}
    \label{tab:softhebb-architecture-depthwise}
    \resizebox{\textwidth}{!}{%
    \setlength{\tabcolsep}{6pt}
    \renewcommand{\arraystretch}{1.2}
    \begin{tabular}{|l|l|l|c|c|c|l|}
    \hline
    \textbf{Layer} & \textbf{Type} & \textbf{Output Shape} & \textbf{Kernel} & \textbf{Stride} & \textbf{Padding} & \textbf{Activation} \\
    \hline
    Input & -- & (3, 32, 32) & -- & -- & -- & -- \\
    \hline
    1 & BatchNorm2d & (3, 32, 32) & -- & -- & -- & -- \\
     & HebbianConv2d & (96, 32, 32) & 5x5 & 1 & 2 & -- \\
     & Triangle & (96, 32, 32) & -- & -- & -- & power=0.7 \\
     & MaxPool2d & (96, 16, 16) & 4x4 & 2 & 1 & -- \\
    \hline
    2 & BatchNorm2d & (96, 16, 16) & -- & -- & -- & -- \\
     & HebbianDepthConv2d & (96, 16, 16) & 3x3 & 1 & 1 & -- \\
     & BatchNorm2d & (96, 16, 16) & -- & -- & -- & -- \\
     & HebbianConv2d & (384, 16, 16) & 1x1 & 1 & 0 & -- \\
     & Triangle & (384, 16, 16) & -- & -- & -- & power=1.4 \\
     & MaxPool2d & (384, 8, 8) & 4x4 & 2 & 1 & -- \\
    \hline
    3 & BatchNorm2d & (384, 8, 8) & -- & -- & -- & -- \\
     & HebbianDepthConv2d & (384, 8, 8) & 3x3 & 1 & 1 & -- \\
     & BatchNorm2d & (384, 8, 8) & -- & -- & -- & -- \\
     & HebbianConv2d & (1536, 8, 8) & 1x1 & 1 & 0 & -- \\
     & Triangle & (1536, 8, 8) & -- & -- & -- & power=1.0 \\
     & AvgPool2d & (1536, 4, 4) & 2x2 & 2 & 0 & -- \\
    \hline
    Output & Linear & (10) & -- & -- & -- & -- \\
    \hline
    \end{tabular}%
    }
\end{table}

\begin{table}[htbp]
    \centering
    \caption{Residual CNN Architecture}
    \label{tab:depthwise-residual-architecture}
    \resizebox{\textwidth}{!}{%
    \setlength{\tabcolsep}{6pt}
    \renewcommand{\arraystretch}{1.2}
    \begin{tabular}{|l|l|l|c|c|c|l|}
    \hline
    \textbf{Layer} & \textbf{Type} & \textbf{Output Shape} & \textbf{Kernel} & \textbf{Stride} & \textbf{Padding} & \textbf{Activation} \\
    \hline
    Input & -- & (3, 32, 32) & -- & -- & -- & -- \\
    \hline
    1 & BatchNorm2d & (3, 32, 32) & -- & -- & -- & -- \\
     & HebbianConv2d & (96, 32, 32) & 5x5 & 1 & 2 & Cosine \\
     & Triangle & (96, 32, 32) & -- & -- & -- & power=0.7 \\
     & MaxPool2d & (96, 16, 16) & 4x4 & 2 & 1 & -- \\
    \hline
    2 & HebbianResidualBlock & (384, 16, 16) & -- & -- & -- & power=1.4 \\
     & MaxPool2d & (384, 8, 8) & 4x4 & 2 & 1 & -- \\
    \hline
    3 & HebbianResidualBlock & (1536, 8, 8) & -- & -- & -- & power=1.0 \\
     & AvgPool2d & (1536, 4, 4) & 2x2 & 2 & 0 & -- \\
    \hline
    Output & Linear & (10) & -- & -- & -- & -- \\
    \hline
    \multicolumn{7}{|c|}{\textbf{HebbianResidualBlock Internal Structure}} \\
    \hline
    Main Path & BatchNorm2d & (in\_ch, H, W) & -- & -- & -- & -- \\
     & HebbianConv2d & (hidden\_dim, H, W) & 1x1 & 1 & 0 & Cosine \\
     & Triangle & (hidden\_dim, H, W) & -- & -- & -- & power=act \\
     & BatchNorm2d & (hidden\_dim, H, W) & -- & -- & -- & -- \\
     & HebbianDepthConv2d & (hidden\_dim, H, W) & 3x3 & 1 & 1 & Cosine \\
     & Triangle & (hidden\_dim, H, W) & -- & -- & -- & power=act \\
     & BatchNorm2d & (hidden\_dim, H, W) & -- & -- & -- & -- \\
     & HebbianConv2d & (out\_ch, H, W) & 1x1 & 1 & 0 & Cosine \\
    \hline
    Shortcut & BatchNorm2d* & (in\_ch, H, W) & -- & -- & -- & -- \\
     & HebbianConv2d* & (out\_ch, H, W) & 1x1 & 1 & 0 & Cosine \\
    \hline
     & Add & (out\_ch, H, W) & -- & -- & -- & -- \\
     & Triangle & (out\_ch, H, W) & -- & -- & -- & power=act \\
    \hline
    \multicolumn{7}{l}{* Only if in\_channels $\neq$ out\_channels} \\
    \end{tabular}%
    }
\end{table}

\end{document}